\documentclass{article}

\usepackage{microtype}
\usepackage{graphicx}
\usepackage{subcaption}
\usepackage{booktabs} 

\usepackage{hyperref}

\usepackage[preprint]{neurips_2026}

\usepackage{amsmath}
\usepackage{amssymb}
\usepackage{mathtools}
\usepackage{amsthm}

\usepackage[capitalize,noabbrev]{cleveref}

\theoremstyle{plain}

\theoremstyle{definition}

\theoremstyle{remark}

\usepackage[textsize=tiny]{todonotes}

\usepackage{array}
\usepackage{makecell}
\usepackage{multirow}
\usepackage{tikz}
\usepackage{xcolor}
\usepackage{wrapfig}

\newcommand{\stepcircled}[1]{%
  \tikz[baseline=(char.base)]{
    \node[shape=circle,draw=black,fill=black,inner sep=1pt] (char) {\textcolor{white}{\textbf{#1}}};
  }%
}

\title{ExplainerPFN: Towards tabular foundation models for model-free zero-shot feature importance estimations}

\author{%
  Joao Fonseca \\
  INESC-ID, Lisbon, Portugal \\
  New York University, New York, USA \\
  \texttt{joaofonseca@nyu.edu} \\
  \And
  Julia Stoyanovich \\
  New York University, New York, USA \\
  \texttt{stoyanovich@nyu.edu} \\
}

\begin{document}

\maketitle

\begin{abstract}

Computing the importance of features in supervised classification tasks is critical for model interpretability. Shapley values are a widely used approach for explaining model predictions, but require direct access to the underlying model, an assumption frequently violated in real-world deployments. We investigate whether meaningful feature attributions can be obtained in a \emph{zero-shot} setting, using only the input data distribution and no evaluations of the target model. Because multiple models can produce identical predictions yet yield different Shapley decompositions, the mapping from data to attributions is not uniquely identifiable. We therefore target attributions that are ``true to the data'' rather than ``true to the model'', learning a posterior mean attribution under a meta-training prior. To this end, we introduce ExplainerPFN, a tabular foundation model built on TabPFN, pretrained on synthetic structural causal datasets supervised with exact or near-exact Shapley values, that predicts feature attributions for unseen tabular datasets without model access, gradients, or example explanations.

Our contributions are fourfold: (1) we show that few-shot surrogate explainers achieve high SHAP fidelity with as few as two reference observations; (2) we propose ExplainerPFN, the first zero-shot method for estimating Shapley-value-style feature attributions without access to the underlying model or reference explanations, providing a principled attribution where no existing explainer can be applied; (3) we release an open-source implementation including the full training pipeline and synthetic data generator; and (4) through extensive experiments on real and synthetic datasets, we show that ExplainerPFN achieves performance competitive with few-shot surrogate explainers that rely on 2--10 SHAP examples.
\footnote{We make all the code developed for this project available at \href{https://github.com/joaopfonseca/ExplainerPFN}{https://github.com/joaopfonseca/ExplainerPFN}}.
\end{abstract}

\section{Introduction}

In many real-world applications, machine learning (ML) systems make or assist in making high-stakes decisions, such as loan approvals, hiring, or medical diagnoses. These models are often proprietary and operate as black boxes, leaving users and regulators without insight into how decisions are made. Consider an individual who is denied a loan by a proprietary, black-box ML system. The applicant may reasonably ask which factors contributed most to the decision. Alternatively, stakeholders may wish to ensure that the model does not distinguish based on protected attributes like race or gender. However, access to the underlying model is rarely granted, and feature-level explanations are almost never provided. In such settings, the ability to estimate feature attributions \emph{without} model access becomes crucial for auditing decisions, identifying potential biases, and improving transparency.

This challenge reflects a broader trend in modern ML. Many real-world models operate behind closed APIs for reasons of privacy, intellectual property, or security, limiting direct model access. At the same time, tabular data has received comparatively little attention in foundation model research.  Recent work highlights both the need for methods that can analyze data without direct access to sensitive models~\cite{hodyou2025do} and the potential of large tabular models to transfer knowledge across heterogeneous tabular tasks~\cite{van2024position}.

Recent studies raise questions about the stability and interpretability of existing feature attribution methods. While covariate-based importance measures are statistically well motivated in many settings~\citep{verdinelli2024feature}, Shapley-based explanations can be sensitive to correlations among input features, and even simple changes in data representation can substantially alter SHAP-based attributions~\citep{hwang2025shap}. Rather than being a purely methodological limitation, these sensitivities suggest that the data distribution itself plays a central role in shaping feature attributions. This observation motivates the possibility of learning attribution patterns directly from data, without requiring access to the underlying model.


We highlight a key subtlety: because different models can yield identical predictions 
yet assign different Shapley values to the same features~\citep{kumar2021shapley,
brunet2022implications}, the mapping from data and predictions to attributions is not 
uniquely identifiable. ExplainerPFN therefore does not aim to recover any specific 
model's SHAP values; given only the data distribution and predictions, it infers the 
attribution structure most probable across the data-generating processes seen during 
training. This posterior mean attribution under a meta-training prior reflects the 
distinction between explanations ``true to the model'' and those ``true to the 
data''~\citep{chen2020true}; ExplainerPFN targets the latter, the only coherent 
target when model access is unavailable. This provides a principled, data-driven 
attribution in settings such as proprietary APIs, regulatory audits, or 
limited-query-budget scenarios, where existing post-hoc explainers cannot be 
applied at all.

\textbf{Contributions.} We study zero-shot estimation of Shapley value feature attributions using only the input data distribution and model predictions. We approach this problem through tabular foundation models and in-context learning, and ask whether Shapley values can be estimated in highly resource-constrained settings without access to the model, gradients, or reference SHAP explanations. We introduce ExplainerPFN, a tabular foundation model built on top of TabPFN~\cite{hollmann2023tabpfn} that predicts Shapley values in a fully zero-shot manner. Our main contributions are:

\textbf{\stepcircled{1}} We show that few-shot surrogate explainers can achieve high SHAP fidelity with as few as two reference observations.

\textbf{\stepcircled{2}} We propose the first \emph{zero-shot} method for estimating Shapley-value-style feature attributions, requiring no access to the underlying ML model or precomputed attribution examples.

\textbf{\stepcircled{3}} We provide an open-source implementation of ExplainerPFN, including the training pipeline, pretrained weights, and a synthetic data generator for tabular foundation model pretraining, adapted from the TabPFN synthetic prior-fitting procedure and redesigned for feature attribution learning.

\textbf{\stepcircled{4}}   Through extensive experiments on real and synthetic datasets, we demonstrate that ExplainerPFN achieves performance competitive with few-shot surrogate explainers requiring 2--10 SHAP examples. 

Overall, our results show a promising path to recover meaningful feature attribution patterns \emph{directly} from the data distribution in zero-access scenarios, opening a new research direction in zero-shot explainability using tabular foundation models.

\section{Preliminaries and Background}\label{sec:related-work}

\textbf{Shapley values.} The Shapley value is a concept from cooperative game theory that provides a fair way to distribute the total gains (or costs) among the players based on their individual contributions to the overall outcome~\cite{shapley1953value}. It has been widely adopted in explainable AI (xAI) to attribute feature importance in ML models. 

Let $f : \mathbb{R}^m \to \mathbb{R}$ be a predictive model, $x_i = (x_i^1, \ldots, x_i^m)$ an input instance\footnote{We use ``input instance'' and ``observation'' interchangeably.}, and $M = \{1, \ldots, m\}$ the set of feature indices. Given a dataset $X = \{x_1, \ldots, x_n\}$, where $x_i^j$ denotes the value of the $j$-ith feature for instance $i$, the Shapley value $\phi_i^j$ is computed as follows:
\begin{equation}\label{eq:shapley}
  \Delta_j(S) \coloneqq \mathbb{E}\left[f(x_i) \mid x_i^{S \cup \{j\}}\right] - 
  \mathbb{E}\left[f(x_i) \mid x_i^S \right]
\end{equation}
where $w(S) \coloneqq |S|!(m - |S| - 1)!/m!$ is the coalition weight, and
$\phi_i^j \coloneqq \sum_{S \subseteq M \setminus \{j\}} w(S) \Delta_j(S)$.
Here, $S$ is a subset of features not including feature $j$, $\mathbb{E}\left[f(x_i) \mid x_i^S \right]$ is the expected model output conditioned on the feature values in $S$, and $|S|$ is the cardinality of $S$. Intuitively, $\Delta_j(S)$ represents the marginal contribution of feature $j$, which is weighted by $w(S)$ to account for all feature coalitions. Finally, $\phi_i^j$ is obtained by summing the weighted marginal contributions over all subsets $S$.

\textbf{SHAP.} SHapley Additive exPlanations (SHAP)~\cite{lundberg2017unified} is a popular game-theoretic framework for local explanations of ML model predictions in the form of feature importance scores derived from Shapley values. For general models (including neural networks), the exact computation of SHAP scores is intractable ($\#P$-hard)~\cite{arenas2023complexity}. Consequently, SHAP (and other Shapley value-based frameworks) typically approximate feature importance either via Monte Carlo sampling over $S$, by limiting the maximum coalition size~\cite{pliatsika2025sharp}, or via model-specific approximations such as TreeSHAP~\cite{lundberg2020local}. These strategies introduce a trade-off between computational cost and fidelity of the explanation, defined by how well the resulting feature importance scores reconstruct the model output. Recent works aim to improve the computational efficiency of SHAP computation in particular settings~\cite{mitchell2022sampling, arenas2023complexity}, typically in polynomial time. Importantly, these methods still require access to the underlying model $f$. Instead, we consider the problem of estimating feature importance when $f$ is not directly accessible. We assume access to $X$ and the corresponding model predictions $\hat{Y} \coloneqq {\hat{y}_1, \ldots, \hat{y}_n}$, where $\hat{y}_i \coloneqq f(x_i)$, and our objective is to approximate $\phi_i^j$.

We focus on SHAP as the target attribution framework for three reasons. First, its model-agnostic, widely used formulation makes it a natural benchmark for zero-shot explainability. Second, its axiomatic properties (e.g., efficiency and symmetry) motivate our post-processing calibration pipeline (Section~\ref{sec:post-processing}). Third, demonstrating zero-shot attribution on a single, well-understood target provides a foundation for extending the approach to other additive attribution methods in future work.

\textbf{Tabular foundation models (TFM).} Formally, a TFM is a pretrained mapping $F \colon \mathcal{X} \times \mathcal{D} \to \mathbb{R}$, where $\mathcal{X} \subseteq \mathbb{R}^m$ denotes the feature space and $\mathcal{D}$ denotes the space of possible context datasets, respectively. At inference time, predictions for $x_i$ are obtained by conditioning on a context dataset $D = \{(x_1, y_1), \ldots, (x_n, y_n)\} \subseteq \mathcal{D}$, i.e., $\hat{y}_i \coloneqq F(x_i \mid D)$. Thus, TFMs leverage in-context learning~\cite{brown2020language}, where $F$ is defined based on priors learned during pretraining (\emph{e.g.}, via synthetic prior-fitting~\cite{hollmann2023tabpfn} or SSL~\cite{ma2025tabdpt}) to make predictions based on $D$, without gradient-based adaptation.

\section{Problem Formulation}\label{sec:problem}

\textbf{Problem statement.} We estimate per-instance feature importance for a 
fixed but inaccessible predictive model $f^b \colon \mathcal{X} \to \mathbb{R}$, 
an instance of $f$ as defined in Section~\ref{sec:related-work}.  Given a dataset of inputs $X = \{x_1, \ldots, x_n\} \subset \mathcal{X}$ and the corresponding model predictions $\hat{Y} = \{\hat{y}_1, \ldots, \hat{y}_n\}$, where $\hat{y}_i = f^b(x_i)$, our goal is to estimate, for each instance $x_i$, a feature importance vector $\phi_i = (\phi_i^1, \ldots, \phi_i^m) \in \mathbb{R}^m$ that approximates the Shapley values of the underlying model. 

We consider problem settings that differ in the degree of access to the model $f^b$ and to reference feature importance scores, ranging from few-shot to fully zero-shot regimes.

\textbf{Few-shot feature importance estimation.}  Let $f^b$ be a base model trained over a distinct labeled dataset $D_{\text{base}}$. Given a small set of instances $X_{\text{train}} \subset \mathcal{X}$ and their corresponding model predictions $\hat{Y}_{\text{train}} = \{f^b(x) \mid x \in X_{\text{train}}\}$ (ground-truth labels are not required), we assume access to a limited collection of reference feature importance vectors $\Phi_{\text{train}} = \{\phi_i \mid x_i \in X_{\text{train}}\}$.

This setting captures two common scenarios: (1) access to $f^b$ is limited, for example via a rate-limited API, so only a small number of reference feature importance vectors can be computed directly, or (2) $f^b$ is entirely inaccessible, but $\Phi_{\text{train}}$ has been precomputed. In either case, the objective is to learn a surrogate model $g$ such that, for an unseen instance $x_i \in X_{\text{test}}$ with predicted output $\hat{y}_i = f^b(x_i)$, the surrogate produces an estimate $g(x_i, \hat{y}_i) \approx \phi_i^{\text{test}}$. 

In this work, we use a tabular foundation model as the surrogate, performing in-context learning over the few-shot reference set. While our primary contribution targets the zero-shot setting, described next, we revisit the few-shot formulation in Section~\ref{sec:experiments} for benchmarking against surrogate explainers with limited reference explanations.

\textbf{Zero-shot feature importance estimation.} Let $f^b$ be a fixed but inaccessible predictive model and suppose that no reference feature importance vectors are available, i.e., $\Phi_{\text{train}} = \varnothing$. Given an unseen dataset $X_{\text{test}} \subset \mathcal{X}$ and the corresponding model predictions $\hat{Y}_{\text{test}} = \{\hat{y}_i \mid x_i \in X_{\text{test}}\}$, the objective is to estimate the feature importance vectors $\Phi_{\text{test}} = \{\phi_i^{\text{test}} \mid x_i \in X_{\text{test}}\}$ without querying $f^b$ and without access to any precomputed explanations.

In this setting, a zero-shot explainer must approximate the mapping $(x_i, \hat{y}_i) \mapsto \phi_i^{\text{test}}$ without observing any triplets $(x_i, \hat{y}_i, \phi_i)$ either at training time or at inference time. The only information available at inference time is the pair $(X_{\text{test}}, \hat{Y}_{\text{test}})$. In Section~\ref{sec:proposed-method} we describe how tabular foundation models can be adapted to address this setting.

\section{ExplainerPFN}\label{sec:proposed-method}

\emph{ExplainerPFN} addresses the zero-shot feature importance estimation problem of Section~\ref{sec:problem}. It adapts the TabPFN framework~\cite{hollmann2023tabpfn} but replaces its predictive objective with a Shapley-value meta-learning objective. ExplainerPFN is trained on a large meta-distribution of synthetic tabular tasks to learn a mapping from input-prediction pairs to per-instance feature attributions, without ever observing real-world explanations. At inference time, ExplainerPFN acts as a conditional, learned attribution generator:
\begin{equation}
  \label{eq:explainerpfn_mapping}
  F_{\text{zs}}: \mathcal{X} \times \mathbb{R} \to \mathbb{R}^m, (x, \hat{y}) \mapsto \hat{\phi},
\end{equation}
Or, equivalently,

\begin{equation}
  \label{eq:explainerpfn_function}
  \hat{\Phi} = F_{\text{zs}}((X, \hat{Y}) \mid (X_\text{ref}, \hat{Y}_\text{ref})) 
  \coloneqq \{\hat{\phi}_1, \ldots, \hat{\phi}_n\}, \quad 
  \hat{\phi}_i = \{\hat{\phi}_i^1, \ldots, \hat{\phi}_i^m\} \in \mathbb{R}^m
\end{equation}
Where $\hat{\Phi} \in \mathbb{R}^{n \times m}$ contains the estimated feature-importance vectors for all $n$ instances in $X$, conditioned on the reference set $(X_\text{ref}, \hat{Y}_\text{ref})$. The reference set may be any dataset representative of the distribution of interest, including $X$ itself or a subset thereof. Crucially, $F_{\text{zs}}$ does not rely on access to $f^b$. This formalism generalizes the few-shot surrogate model $g$ introduced in Section~\ref{sec:problem} to the zero-shot setting.

We note that the mapping $(X, \hat{Y}) \to \hat{\Phi}$ is not uniquely identifiable: different models can yield identical predictions while assigning different Shapley  values~\citep{kumar2021shapley,brunet2022implications}. ExplainerPFN therefore does not aim to recover the SHAP values of any specific $f^b$. Instead, it learns a posterior mean attribution under the meta-training prior---i.e., given the data distribution and predictions, it infers the attribution structure most probable across the data-generating processes and  models seen during pretraining---using SHAP values as training supervision on  synthetic tasks with known ground truth, not as an intermediate quantity recomputed at inference time. This follows the distinction of~\citep{chen2020true} between explanations ``true to the model'' and ``true to the data'' (see Appendix~\ref{app:true-to-data}). In zero-access settings without model queries, ExplainerPFN  provides a principled, data-driven attribution that existing post-hoc methods cannot produce.

\subsection{Tabular Transformer Encoder}

To implement the zero-shot mapping defined in Equations \eqref{eq:explainerpfn_mapping} and \eqref{eq:explainerpfn_function}, ExplainerPFN uses a transformer encoder~\cite{vaswani2017attention} adapted for tabular data, similar to TabPFN. Each pair $(x_i, \hat{y}_i)$ is encoded as a token, producing a $d$-dimensional embedding that integrates feature values, predicted output, and positional information. Rows interact through multiple layers of bidirectional self-attention, capturing intra- and inter-instance dependencies relevant for feature attribution. The embedding corresponding to the instance being explained after the final layer is used to produce the estimated attribution $\hat{\phi}_i^j$ for each feature.

For each feature $j$, inputs are reordered so that $\hat{y}_i$ and $x^j_i$ occupy the first two columns before encoding, enabling a consistent mapping from position to attribution. Unlike TabPFN, which masks test labels during training and inference to simulate the prediction task, ExplainerPFN uses the full $(X, \hat{Y})$ pairs for all examples; $F_\text{zs}$ accepts $X$ and $X_\text{ref}$ of arbitrary size, and $\Phi$ is never used as input.

%

\subsection{Generating Training Data}\label{sec:training-data}

Training ExplainerPFN requires ground-truth Shapley values, which are  computationally infeasible to obtain at scale for real data. We therefore  generate all training tasks synthetically, following the prior-fitting  paradigm of~\citet{hollmann2025accurate} adapted for feature attribution. Each meta-training task is generated as follows (see Appendix~\ref{sec:appendix_data_generation} for details).


\textbf{\stepcircled{1}} \emph{Sampling a structural causal model.}
We sample a random directed acyclic graph (DAG) representing a structural causal model (SCM), which defines a synthetic tabular data-generating process. Feature values are generated by propagating noise through the graph to obtain a synthetic dataset $D$, from which we select $m \leq s$ nodes at random to form the input features $X$ and one node to form a binary target $Y$, and discarding the remaining nodes. This construction yields a dataset consistent with the binary prediction setting described in Section~\ref{sec:related-work}. Figure~\ref{fig:dag_ground_truth} shows an example  DAG.

\textbf{\stepcircled{2}} \emph{Training a base model.} Next, we define and train $f^b$. We use a simple feed-forward neural network with one hidden layer of size 100 and ReLU activations. The model is trained on a subset of $D$ for 2000 epochs using the Adam optimizer with initial learning rate $\eta_0 = 0.0001$ and scheduled decay $\eta_t = \frac{\eta_0}{\sqrt{t+1}}$. This yields model predictions $\hat{Y} = f^b(X)$.

\textbf{\stepcircled{3}} \emph{Computing ground-truth Shapley values.}
Finally, we compute $\phi_i = (\phi_i^1, \ldots, \phi_i^m)$ for model $f^b$ over all observations in $X$ using SHAP (see Section~\ref{sec:related-work}). We apply a hybrid approach that uses exact computation for low-dimensional feature sets and permutation-based SHAP estimation otherwise, yielding a set of feature attributions $\Phi = \{\phi_1, \ldots, \phi_n\}$ while keeping computation tractable.

This procedure produces triplets $(X, \hat{Y}, \Phi)$, required for training ExplainerPFN, sampled from a meta-distribution of synthetic tabular tasks. Because the data-generating processes, base model parameters, and target features vary across tasks, ExplainerPFN learns a distribution-level prior over the relationship between $(x, \hat{y})$ pairs and $\phi$.

Synthetic training serves three purposes: it prevents data leakage from evaluation datasets; it avoids the prohibitive cost of computing exact SHAP for thousands of real tasks; and it follows the prior-fitting 
paradigm~\citep{hollmann2023tabpfn,hollmann2025accurate}, where the model learns an inference algorithm over a broad prior rather than fitting to static data. The only exception is the DAG-structure experiment in Section~\ref{sec:exp_dag_structure}, conducted on synthetic data to assess structural recovery.

To prevent overfitting to a single explanation-generating process, we diversify meta-training labels in two ways. First, we vary the initialization of non-parametric base models across tasks, exposing ExplainerPFN to a broader range of attribution patterns arising from model instances with similar predictive behavior~\citep{black2022model,brunet2022implications}. Second, we compute ground-truth labels using multiple Shapley approximation strategies and random seeds rather than a single fixed estimator, reducing label bias from any particular approximation method. Together, these choices encourage ExplainerPFN to learn attribution structure that is stable across plausible model- and estimation-generating processes.


\subsection{Training Objective}\label{sec:training-objective}

ExplainerPFN follows the probabilistic training setting of TabPFN but adapts it to the feature-attribution setting. Rather than predicting a scalar output, the model produces a posterior predictive distribution over $K$ discretized ``buckets'' representing possible values of the standardized attribution:
\begin{equation}
  p(\tilde{\phi}_i^j = b_k \mid x_i, \hat{y}_i, X_{\text{ref}}, \hat{Y}_{\text{ref}}), \quad k = 1, \ldots, K,
\end{equation}
The final point estimate used at inference time corresponds to the expectation of this distribution,
\begin{equation}
  \hat{\phi}_i^j = \sum_{k=1}^K b_k \cdot p(\tilde{\phi}_i^j = b_k \mid x_i, \hat{y}_i, X_{\text{ref}}, \hat{Y}_{\text{ref}}).
\end{equation}
Training minimizes the Negative Log Predictive Density: 
\begin{equation}
  \text{NLPD} = - \sum_{i=1}^n \sum_{j=1}^m \log p(\tilde{\phi}_i^j = b_{k^*} \mid x_i, \hat{y}_i, X_{\text{ref}}, \hat{Y}_{\text{ref}}),
\end{equation}
where $b_{k^*}$ is the bucket containing the true standardized attribution $\tilde{\phi}_i^j$. NLPD is well established in probabilistic modeling~\citep{candela2003propagation, 
yu2021leveraging}, encouraging accurate predictions and well-calibrated uncertainty 
estimates. Optimization uses Adam with cosine-annealed learning rates sampled from 
$[10^{-7}, 10^{-4}]$, selecting the best-performing schedule based on final training loss.

Since raw Shapley values often have small magnitudes and vary in scale across tasks, we normalize the targets using $\tilde{\phi}_i^j = \frac{\phi_i^j - \mu}{\sigma}$, where $\mu$ and $\sigma$ denote the mean and standard deviation of all Shapley values within a meta-training task. This preserves relative scaling between features while ensuring numerical stability during optimization.

\subsection{Improving Attribution Consistency via Shapley Properties}
\label{sec:post-processing}

Although ExplainerPFN is trained to predict Shapley values directly, its raw outputs need not satisfy Shapley axioms (symmetry and efficiency~\cite{shapley1953value}) exactly. We apply a three-step post-processing pipeline to improve between-feature comparability and explanation fidelity, using only the predicted attributions and model outputs---no access to the underlying model is required. The three steps 
are: (1)~\emph{mean-zero re-centering}, enforcing efficiency in expectation by re-centering attributions to zero mean across the dataset; (2)~\emph{variance rescaling}, normalizing per-feature attribution variance under the independence assumption of Linear-SHAP~\cite{lundberg2017unified}; and (3)~\emph{instance-level efficiency correction}, ensuring each instance's attributions sum to the model output minus the baseline. See Appendix~\ref{app:post-processing} for full derivations.

\section{Experiments}\label{sec:experiments}

We evaluate ExplainerPFN under highly limited or absent access to feature attributions. Because true zero-shot Shapley estimators are unavailable, we compare ExplainerPFN’s predictions to SHAP values computed from standard ML models trained with few supervision points, which serve as a practical proxy in low-data regimes. 

\textbf{Datasets.} We evaluate ExplainerPFN on 11 tabular datasets from the UCI repository, all unseen during training, and include a real-world case study using the ACS Public Coverage dataset~\cite{ding2021retiring} (Section~\ref{sec:case_study}). Additional dataset details are provided in Table~\ref{tab:dataset_summary} in Appendix~\ref{sec:appendix_datasets}. These datasets serve as evaluation benchmarks for zero-shot transfer: the foundation model's purpose is to learn generalizable attribution patterns from synthetic meta-training without any per-dataset training or model access.

\textbf{Base predictors.} For each dataset we use two base predictors, $f^b$, each preceded by a standard scaler (\emph{i.e.}, Z-score normalization): a Multi-Layer Perceptron (MLP) with two hidden layers of size 12 and ReLU activations, and a Random Forest (RF) classifier with 500 estimators. Both provide non-linear decision boundaries representative of practical classifiers.

\textbf{Feature importance predictors.} We compare ExplainerPFN against few-shot surrogate explainers trained using small sets of SHAP explanations: a TabPFN regressor, an MLP regressor, and an RF regressor, each trained with 2–10 SHAP examples per dataset. All TabPFN and ExplainerPFN results use a single forward pass without ensembling.

\textbf{Metrics.} We use exact SHAP values from the base predictors as reference. 
Our primary metric is Pearson correlation between predicted and reference attributions, 
capturing agreement in relative magnitude and sign. We also report Spearman rank 
correlation for scale-independent ordering agreement. We report bootstrap 95\% 
confidence intervals (50 resamples with replacement) for all metrics; full results 
are in Appendix~\ref{sec:appendix_full_tables}.

\textbf{Hardware.} Experiments were conducted on a machine with a 14-core Intel Xeon Platinum 8268 (2.90GHz) CPU, 32GB RAM, and an NVIDIA V100 GPU with 32GB VRAM.

\subsection{Few-shot Feature Importance Estimation}\label{sec:exp_few_shot}

We first examine the performance of few-shot surrogate explainers trained using a small number of SHAP explanations from the target model. Tables~\ref{tab:results_corr_across_datasets_rfr} and~\ref{tab:results_corr_across_datasets_mlp} report results for several supervised surrogates trained with 2--10 reference explanations. Notably, few-shot methods, particularly TabPFN-based surrogates, often achieve high correlation with ground-truth SHAP values using as few as two reference observations, with performance improving rapidly as additional supervision becomes available.  For example, when using a random forest as the base predictor, TabPFN trained with four reference explanations already achieves correlations of 0.667 on BA and 0.846 on EC, and reaches correlations comparable to or exceeding other surrogates trained with substantially more supervision as the number of reference explanations increases (0.903 on BA and 0.963 on EC with ten references). 

\begin{table}
    \centering
    \caption{Pearson correlation between true Shapley values computed based on a RFR (as the base predictor) and estimated Shapley values using ExplainerPFN (zero-shot predictions), as well as a MLP, a RF and TabPFN in different few-shot settings with varying numbers of training samples. Results are reported across a diverse set of UCI datasets.}
    \label{tab:results_corr_across_datasets_rfr}
    \vspace{0.5em}
    \footnotesize
    \resizebox{\columnwidth}{!}{\begin{tabular}{ccccccccccccc}
\toprule
Method & Samples & BA & AB & AN & CC & EC & FL & GC & HE & HD & HP & TH \\
\midrule
\multirow[c]{1}{*}{EPFN} & 0 & \underline{0.897} & 0.177 & 0.443 & \underline{0.479} & 0.397 & 0.098 & 0.160 & \textbf{0.641} & \textbf{0.580} & \textbf{0.643} & 0.317 \\
\midrule
\multirow[c]{5}{*}{MLP} & 2 & -0.058 & 0.142 & 0.001 & -0.013 & 0.040 & 0.205 & 0.150 & -0.073 & 0.230 & -0.083 & 0.053 \\
 & 4 & 0.019 & 0.130 & 0.093 & 0.079 & 0.218 & 0.074 & 0.159 & -0.140 & 0.070 & -0.081 & 0.067 \\
 & 6 & 0.065 & -0.061 & 0.091 & 0.075 & 0.320 & 0.094 & 0.078 & -0.132 & 0.136 & -0.078 & 0.065 \\
 & 8 & 0.656 & 0.023 & 0.088 & 0.090 & 0.316 & 0.129 & 0.117 & -0.112 & 0.204 & -0.060 & 0.070 \\
 & 10 & 0.599 & -0.065 & 0.091 & 0.065 & 0.445 & 0.066 & 0.119 & -0.091 & 0.088 & -0.057 & 0.062 \\
\midrule
\multirow[c]{5}{*}{RF} & 2 & 0.376 & 0.037 & 0.041 & 0.134 & 0.304 & 0.121 & 0.050 & -0.020 & 0.220 & 0.413 & 0.035 \\
 & 4 & 0.556 & 0.093 & 0.176 & 0.251 & 0.520 & 0.427 & 0.182 & 0.260 & 0.287 & 0.434 & 0.095 \\
 & 6 & 0.562 & \textbf{0.377} & 0.272 & 0.332 & 0.767 & 0.561 & 0.283 & 0.359 & 0.315 & 0.340 & 0.310 \\
 & 8 & 0.767 & 0.274 & \textbf{0.569} & 0.351 & 0.738 & \underline{0.577} & \underline{0.291} & 0.338 & 0.370 & 0.420 & 0.351 \\
 & 10 & 0.879 & 0.285 & \underline{0.556} & 0.375 & \textbf{0.962} & \textbf{0.596} & \textbf{0.327} & 0.400 & 0.434 & 0.431 & \underline{0.594} \\
\midrule
\multirow[c]{5}{*}{TabPFN} & 2 & 0.109 & 0.045 & -0.120 & 0.080 & 0.190 & 0.104 & 0.065 & -0.117 & -0.056 & -0.047 & 0.034 \\
 & 4 & 0.668 & 0.040 & 0.226 & \textbf{0.527} & 0.845 & 0.540 & 0.156 & 0.405 & 0.181 & 0.483 & 0.139 \\
 & 6 & 0.738 & \underline{0.296} & 0.280 & 0.297 & 0.942 & 0.491 & 0.133 & 0.388 & 0.345 & 0.486 & 0.326 \\
 & 8 & 0.773 & 0.221 & 0.510 & 0.345 & 0.952 & 0.341 & 0.245 & 0.445 & 0.388 & \underline{0.546} & 0.404 \\
 & 10 & \textbf{0.903} & 0.232 & 0.517 & 0.446 & \underline{0.962} & 0.531 & 0.247 & \underline{0.448} & \underline{0.459} & 0.531 & \textbf{0.732} \\
\bottomrule
\end{tabular}}
\end{table}

These results demonstrate that high-fidelity feature attributions are attainable even under extremely limited access to explanation supervision. Spearman rank correlation results for the RF base predictor are provided in Appendix~\ref{sec:appendix_experimental_results}. This few-shot regime provides a useful reference point for evaluating zero-shot methods, which we consider next.

\subsection{Zero-shot Feature Importance Estimation}\label{sec:exp_zero_shot}

\begin{figure}[b!]
    \centering
    \begin{subfigure}{0.4\linewidth}
        \centering
        \includegraphics[width=\linewidth]{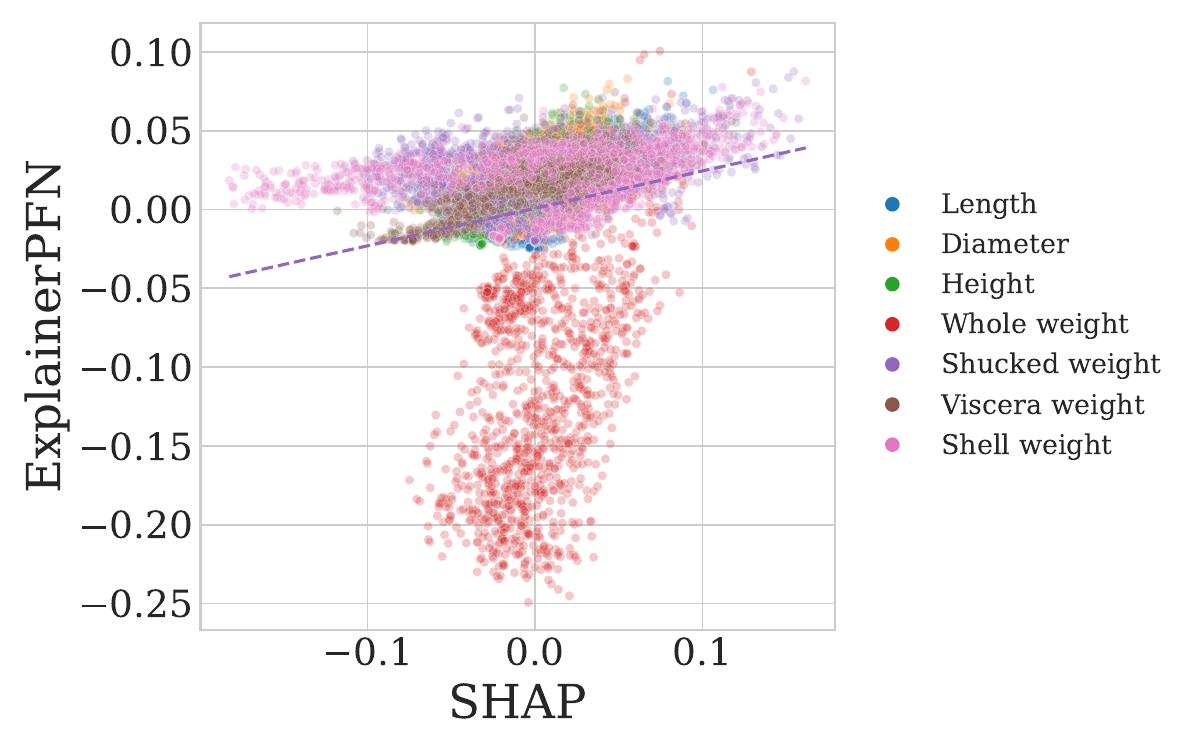}
        \caption{}
        \label{fig:ab_overall_comparison}
    \end{subfigure}
    \begin{subfigure}{0.4\linewidth}
        \centering
        \includegraphics[width=\linewidth]{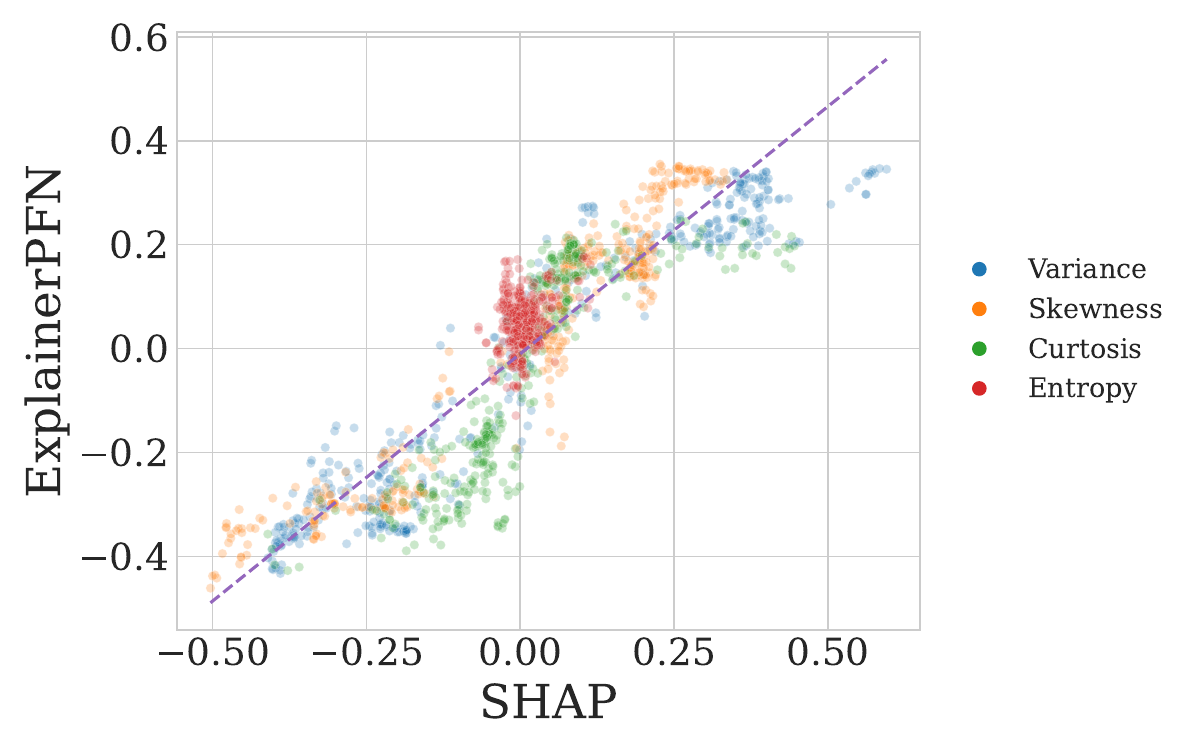}
        \caption{}
        \label{fig:ba_overall_comparison}
    \end{subfigure}
    \caption{Comparison between ExplainerPFN and SHAP feature-importance attributions across the worst and best performing cases for ExplainerPFN, using the RF as base model: \textbf{(a)} Abalone (AB) and \textbf{(b)} Banknote Authentication (BA).}
    \label{fig:overall_comparison}
\end{figure}

We evaluate ExplainerPFN in a true zero-shot setting, where no access is available to the base predictor $f^b$ or to reference SHAP explanations. We compare ExplainerPFN against three few-shot surrogate explainers trained with 2--10 SHAP explanations from the base predictor. This comparison is intentionally stringent: while ExplainerPFN operates without supervision, the baselines benefit from direct access to exact Shapley values from the target model.

Table~\ref{tab:results_corr_across_datasets_rfr} reports results using a RF as $f^b$. ExplainerPFN achieves competitive performance across most datasets, attaining the highest or second highest correlation with SHAP on 5 of 11 datasets and second-best performance on BA. These results indicate that meaningful attribution structure can be recovered from the data distribution alone, even without model access. ExplainerPFN also consistently outperforms MLP-based surrogate explainers across nearly all configurations.
%
We extend this analysis to MLP base predictors in Appendix~\ref{sec:appendix_experimental_results}, both for completeness and to assess robustness to model multiplicity~\citep{black2022model,brunet2022implications}.
To understand this variability, we characterize the UCI datasets along dimensionality and pairwise feature interactions via normalized mutual information, correlating each with zero-shot performance (Appendix~\ref{sec:appendix_characterization_scalability}). Dimensionality emerges as the dominant predictor of degradation, while interaction depth shows only a weak negative association.

\textbf{Best- and worst-case zero-shot behavior.} We analyze ExplainerPFN’s behavior in representative best- and worst-case scenarios identified in Table~\ref{tab:results_jaccard_across_datasets_rf}, using RF as the base model. Figure~\ref{fig:ab_overall_comparison} shows results on the AB dataset, where ExplainerPFN performs poorly. While the method captures the relative ordering of feature importance, it fails to accurately recover absolute magnitudes, leading to overestimation of a negatively contributing feature and a distorted attribution distribution after post-processing.
In contrast, Figure~\ref{fig:ba_overall_comparison} shows results on the BA dataset, where ExplainerPFN closely aligns with SHAP both in relative ordering and absolute magnitude. This comparison highlights both the current limitations of zero-shot estimation, particularly in between-feature scaling, and its potential to produce accurate explanations when data characteristics are favorable.

\textbf{Effect of Shapley-consistent post-processing.}  We analyze the impact of the post-processing steps in Section~\ref{sec:post-processing} on zero-shot performance using synthetic datasets with ground-truth Shapley values, shown in Figure~\ref{fig:ablation_combined}, and extend this discussion in Appendix~\ref{sec:appendix_ablation}. While raw ExplainerPFN outputs capture meaningful attribution structure, applying the full post-processing pipeline consistently improves alignment with ground-truth Shapley values compared to raw outputs or partial corrections. The mean-zero and variance-based scaling steps help align the scale and offset of predicted attributions, which in turn enhances the effectiveness of the instance-level efficiency correction. However, we also find that ExplainerPFN outputs tend to degrade as feature dimensionality increases.

\begin{figure}
    \centering
    \begin{subfigure}{0.35\linewidth}
        \centering
        \includegraphics[width=\linewidth]{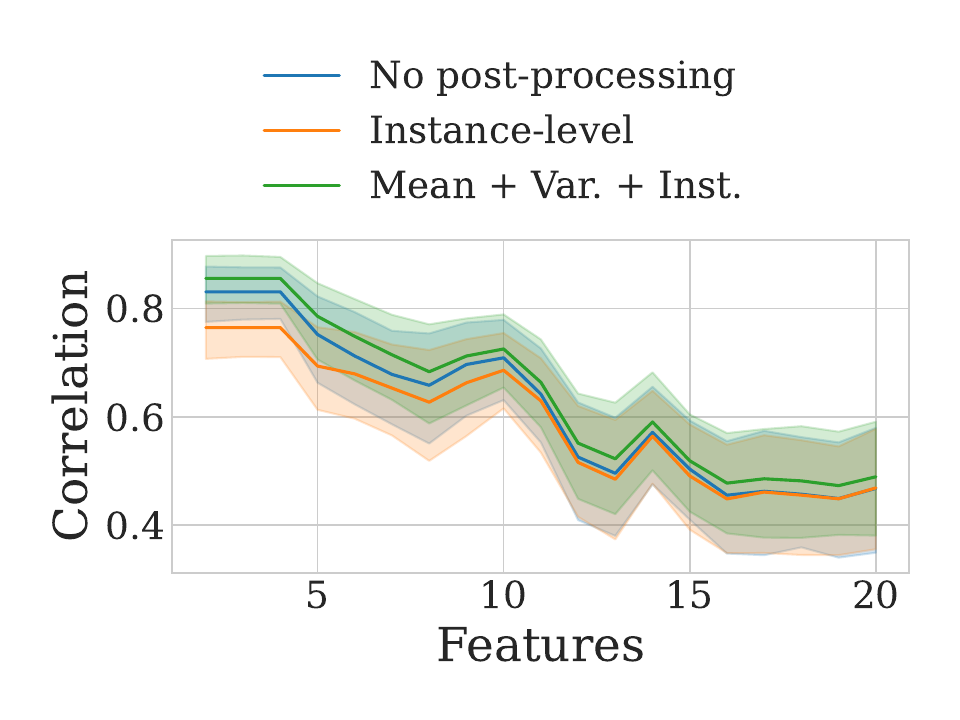}
        \caption{}
        \label{fig:ablation_features}
    \end{subfigure}
    \hfill
    \begin{subfigure}{0.35\linewidth}
        \centering
        \includegraphics[width=\linewidth]{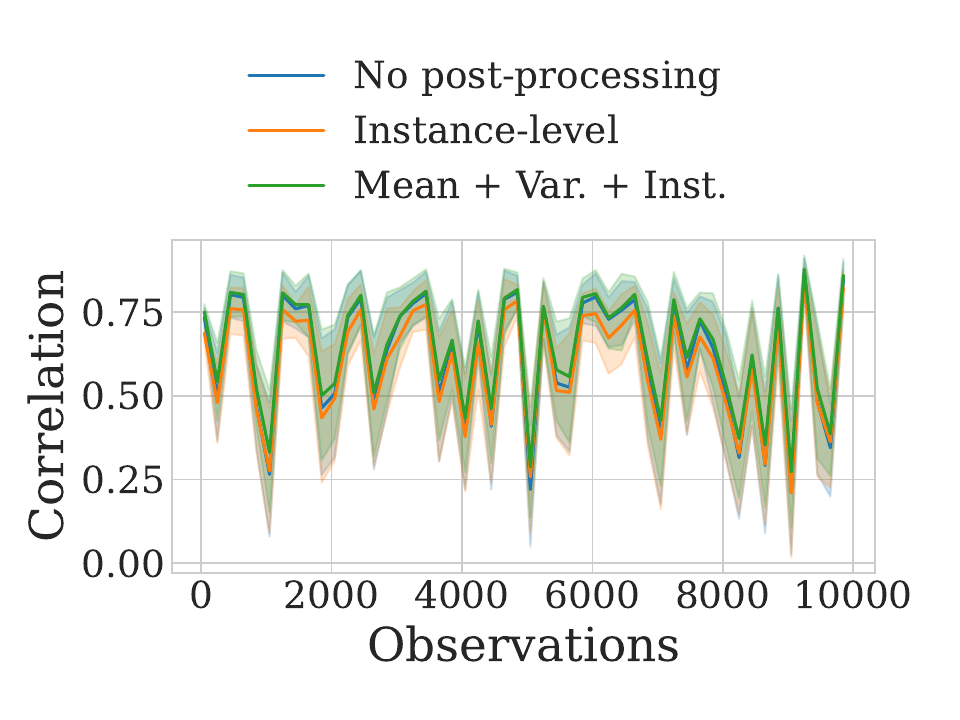}
        \caption{}
        \label{fig:ablation_observations}
    \end{subfigure}
    \caption{Pearson correlation analysis between ExplainerPFN and SHAP feature importance attribution on synthetic datasets using different correction steps. \textbf{(a)} As a function of the number of features. \textbf{(b)} As a function of the number of observations. On average, no post-processing (blue) outperforms instance-level efficiency correction (orange); both are consistently outperformed by applying all correction steps (green; mean-zero attributions, variance decomposition, and instance-level efficiency).}
    \label{fig:ablation_combined}
\end{figure}

\begin{figure}[b!]
    \centering
    \includegraphics[width=0.9\linewidth]{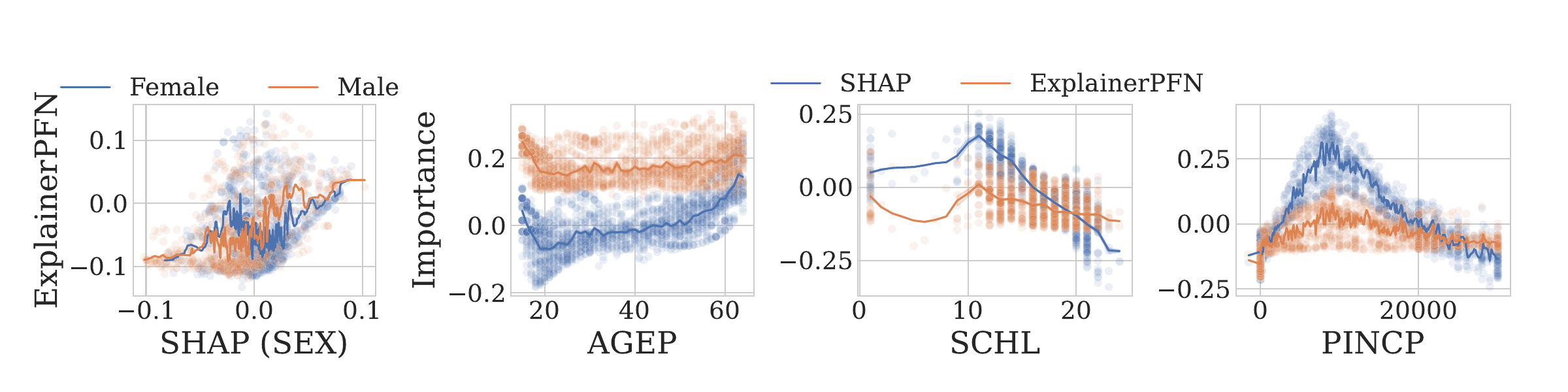}
    \caption{Per-feature comparison between ExplainerPFN and SHAP's feature importance attributions for predicting health insurance coverage using the ACS Public Coverage dataset.}
    \label{fig:acs_feature_importance}
\end{figure}

\textbf{Robustness across base predictors.} We assess robustness to model multiplicity~\citep{black2022model,brunet2022implications} by evaluating zero-shot performance across different base predictors. We repeat the zero-shot experiments using a multi-layer perceptron as the base model instead of a random forest (Appendix~\ref{sec:appendix_zero_shot_mlp}). Despite differences in model architecture and the resulting SHAP explanations, ExplainerPFN exhibits similar performance trends across base predictors, maintaining competitive correlation with ground-truth Shapley values on multiple datasets. This consistency indicates that ExplainerPFN does not overfit to attributions from a single model class, and that the training-time diversification of base model initializations described in Section~\ref{sec:training-data} contributes to robustness under model multiplicity. Additional analyses are provided in
Appendix~\ref{sec:appendix_experimental_results}.

\subsection{Case study: ACS Public Coverage}\label{sec:case_study}

\begin{wrapfigure}{r}{0.45\linewidth}
    \centering
    \includegraphics[width=\linewidth]{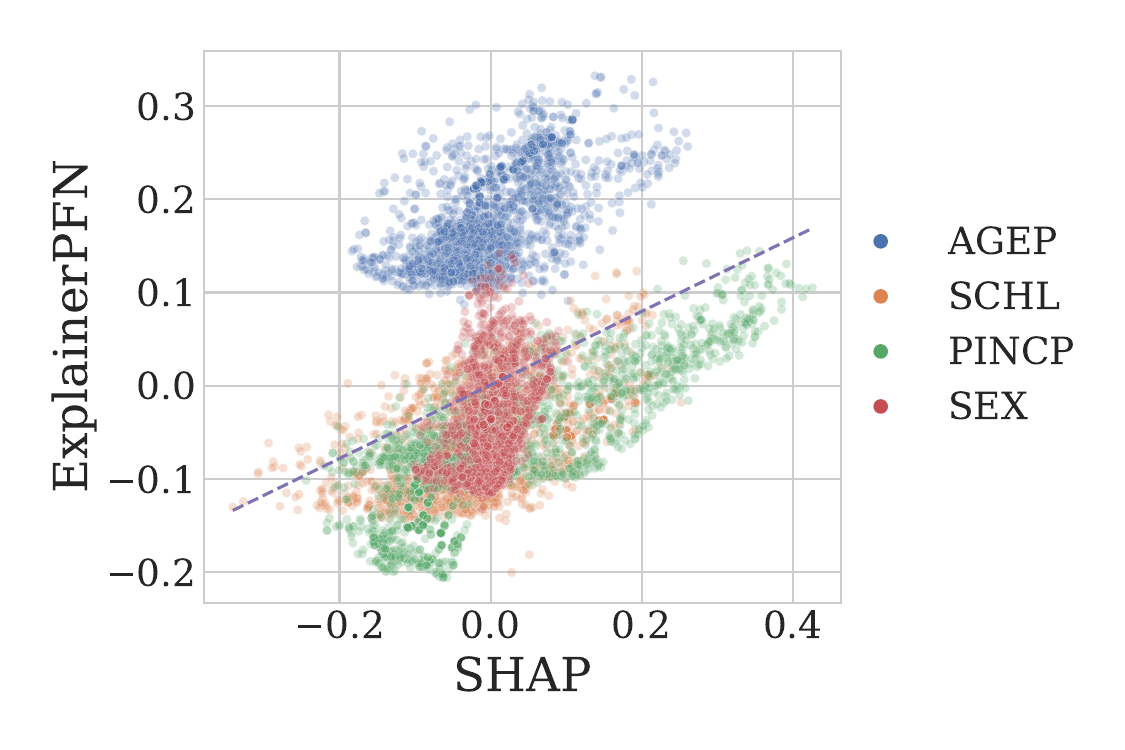}
    \caption{ExplainerPFN vs SHAP feature importance for predicting health insurance coverage in the ACS Public Coverage dataset.}
    \label{fig:acs_overall_comparison}
\end{wrapfigure} 
We evaluate ExplainerPFN on ACS Public Coverage from the 2018 1-Year American Community Survey for Alabama~\cite{ding2021retiring}, accessed via the \texttt{folktables} library. The task is to predict public health insurance coverage (PCOV) using four features: age (AGEP), education (SCHL), income (PINCP), and sex (SEX). We use stratified training and test splits of 2,000 samples each.  This is a challenging case for ExplainerPFN, with a Pearson correlation of 0.32 relative to SHAP, placing it at the lower end of the performance range observed in Section~\ref{sec:exp_zero_shot}. Examining such cases helps clarify the method’s limitations.  Figure~\ref{fig:acs_feature_importance} compares per-feature attributions from ExplainerPFN and SHAP. Although the methods differ in scale, ExplainerPFN recovers consistent directional patterns. Figure~\ref{fig:acs_overall_comparison} further compares SHAP and ExplainerPFN values across all features and instances: SCHL and PINCP show strong linear agreement, indicating consistent importance and effect direction, while SEX exhibits weaker but still positive alignment. AGEP shows the largest discrepancy, with consistent sign but an overestimation of  contribution magnitudes by ExplainerPFN relative to SHAP. Extended per-instance waterfall comparisons are provided in Appendix~\ref{sec:appendix_acs_case_study}.


\section{Conclusions and Limitations}\label{sec:conc}

We introduced the task of zero-shot, model-free feature importance estimation and proposed ExplainerPFN, a tabular foundation model that predicts Shapley-value-style feature attributions from input-prediction pairs without access to the underlying model or reference explanations. Because the mapping from data and predictions to attributions is non-identifiable, ExplainerPFN learns a posterior mean attribution under a 
meta-training prior, targeting ``true to the data'' rather than ``true to the model'' explanations~\citep{chen2020true}. Across a range of real and synthetic datasets, ExplainerPFN achieves competitive performance relative to few-shot surrogate explainers while providing substantial computational savings when model access is limited or SHAP computation is costly. Our results suggest that meaningful attribution structure can be recovered from the data distribution alone in zero-access settings.

\textbf{Limitations.} Our experiments also surface limitations of the current approach. \emph{Cross-feature calibration.} ExplainerPFN recovers relative ordering and  directional effects but systematically misestimates absolute magnitudes across features,  limiting its usefulness in settings where between-feature scale comparisons matter. \emph{Dimensionality and interactions.} Zero-shot quality degrades as feature 
dimensionality grows and interactions become more complex; the current architecture does not scale reliably to high-dimensional regimes. \emph{Synthetic-to-real transfer.} Training exclusively on synthetic tasks prevents leakage but introduces a distribution gap: the synthetic generator may not capture key structural properties of real tabular data, particularly for mixed-type tables and tasks with strong distribution shift.
\emph{Scope.} ExplainerPFN currently targets SHAP specifically; it is unclear whether the meta-training prior transfers to other additive attribution frameworks such as integrated gradients, LIME, or causal attributions.

Despite these limitations, our results demonstrate that meaningful attribution  structure can be recovered from data alone, establishing zero-shot explainability as a viable direction for settings where model access is unavailable.
\begin{ack}
This work was supported in part by US National Science Foundation (NSF) Awards No. 2312930, 2326193. This work was also supported in part by the NYU-KAIST Partnership and by the Institute of Information \& Communications Technology Planning \& Evaluation (IITP) with a grant funded by the Ministry of Science and ICT (MSIT) of the Republic of Korea in connection with the Global AI Frontier Lab International Collaborative Research. (No. RS-2024-00469482 \& RS-2024-00509258). This work was also supported in part by the European Union's Horizon Europe research and innovation program under Grant Agreement No. 101189689, and by Funda\c{c}\~{a}o para a Ci\^{e}ncia e a Tecnologia, I.P. (FCT) under projects UID/50021/2025 (DOI: \url{https://doi.org/10.54499/UID/50021/2025}) and UID/PRR/50021/2025 (DOI: \url{https://doi.org/10.54499/UID/PRR/50021/2025}).
\end{ack}

\bibliography{references}
\bibliographystyle{plainnat}

\newpage
\appendix
\section*{Impact Statement}

We introduce the task of zero-shot, model-free feature importance estimation and propose a Tabular Foundation Model that produces Shapley-like explanations without requiring access to the underlying predictive model. By enabling explanations in settings where the model cannot be inspected directly, this work aims to expand the practical reach of explainable AI. Potential positive impacts include greater transparency in automated decision systems, facilitating preliminary audits for fairness and bias when internal models are inaccessible, and reducing the computational burden associated with traditional Shapley value computation in large-scale applications.  We view these explanations primarily as a triage tool under realistic access constraints, not as a substitute for model-access explanation methods when those are available.

However, we also acknowledge important limitations and potential risks. Because ExplainerPFN operates without model access, its attributions are approximate and may not reflect the true causal behavior of specific underlying models. Users may misinterpret these explanations if they assume they are exact, leading to overconfidence in conclusions about model behavior. The method may also yield less reliable results in high-dimensional settings or in domains where the synthetic pretraining distribution fails to capture critical real-world structure, potentially limiting its applicability without further adaptation. In high-stakes settings, approximate attributions could also be misused as evidence of compliance or fairness without appropriate validation.

Broader deployment of zero-shot explainability techniques should be accompanied by efforts to understand and document their appropriate use cases and limitations. Future research could explore formal guarantees, robustness under distribution shift, and best practices for integrating such tools into responsible ML pipelines. When possible, we recommend validating key findings against model-access explanations and reporting sensitivity indicators (e.g., stability across reference sets or seeds) to reduce the risk of over-interpretation.  This method does not solve all challenges of transparent decision-making, but we hope it encourages further investigation into explanation methods that are practical and informative under realistic access constraints.

To promote future research in this challenging, societally beneficial direction, we release all the artifacts (code, models, data, benchmarks, etc.) produced during this work. 

\section{Related Work}\label{sec:literature-review}

\textbf{Transformers for tabular data.} Earlier works, such as SAINT~\cite{somepalli2021saint}, TabTransformer~\cite{huang2020tabtransformer}, and FT-Transformer~\cite{gorishniy2021revisiting}, have adapted the transformer architecture for supervised learning on tabular data. These models typically employ self-attention mechanisms to attend to both individual features within each observation and across different observations in the dataset. These models are designed for supervised learning on a single dataset and require fine-tuning on new datasets. 

\textbf{Tabular foundation models.} Recent advances in foundation models for tabular data have demonstrated the effectiveness of large pretrained architectures for supervised learning tasks on tabular data. Early work in this direction includes TabPFN~\cite{hollmann2023tabpfn}, which frames tabular prediction as a sequence modeling problem and trains a transformer to emulate the Bayesian inference procedure of a neural process. A refined and more accurate iteration of TabPFN~\cite{hollmann2025accurate} extends the original formulation in several ways, particularly support for regression tasks and greater robustness to unimportant features and outliers. Other recent approaches such as LaTable~\cite{van2024latable} adopt a diffusion-based approach that incorporates dataset descriptions and column names to improve its generative accuracy. More recently, TabDPT~\cite{ma2025tabdpt} achieved strong downstream performance on unseen regression and classification benchmarks via self-supervised learning (SSL).

\textbf{Diffusion models for tabular data generation.} TabDiff~\cite{shi2025tabdiff} is a diffusion-based model for generating tabular data. It uses a transformer-based architecture to model the denoising process. However, TabDiff requires retraining on each new dataset, making it unsuitable for zero-shot learning scenarios. Similarly, MTabGen~\cite{villaizan2025diffusion} is a diffusion model for data imputation and synthetic data generation on tabular data, which generalizes TabDDPM~\cite{kotelnikov2023tabddpm}. Other methods, such as TabCSDI~\cite{zheng2022diffusion}, are aimed towards handling missing value imputation in tabular data by supporting both numerical and categorical variables. However, to the best of our knowledge, none of these diffusion models are designed for in-context learning.

\textbf{Network-based feature importance estimation.} \citet{vskrlj2020feature} propose a Self-Attention Network that generates feature importance scores alongside predictions. However, it must be trained over each target dataset to predict the target feature and the explanations correspond to the attention weights over features. In practice, this approach explains a secondary model’s internal attention mechanism, whose predictions and explanation may not align with feature importance of the base model's predictions. More recently, \citet{cheng_unifying_2025} proposed ShapTST, a framework that integrates Shapley value-based explanation generation directly into time-series transformer training. This enables models to output both predictions and feature importance explanations in a single forward pass. However, ShapTST is specifically designed for time-series data, and requires retraining for each new dataset, along with computation of Shapley values (or approximations thereof) during training. To the best of our knowledge, our work is the first to propose a tabular foundation model that generates feature importance explanations in a zero-shot manner without retraining on new datasets.

\textbf{Meta-learning for tabular data.} Meta-learning exposes models to many small tabular tasks, enabling generalization to new tasks~\cite{thrun1998learning, schmidhuber1987evolutionary, naik1992meta}. MAML~\cite{finn2017model} is a popular model-agnostic training algorithm (a meta-learning procedure) that learns a good initialization for the underlying model. However, this approach requires fine-tuning on new tasks, making it unsuitable for zero-shot learning scenarios. Prototypical Networks~\cite{snell2017prototypical} learn an embedding space where each class is represented by the mean (prototype) of its few labeled examples, and classification is done by nearest prototype using a distance-based softmax. This method is extended for zero-shot learning, but for both the few-shot and zero-shot setting, this method still requires gradient updates on new tasks and is exclusive for classification. Recent work has extended this approach to regression tasks~\cite{gurungprotopairnet}. Task2Vec~\cite{achille2019task2vec} represents each supervised visual classification task as a fixed-length vector. It allows for measuring task similarity and selecting the best pretrained model for a new task. However, this method assumes access to different pretrained models, which are scarce for tabular data.

\textbf{Zero-shot Learning on tabular data.} TabLLM explores the application of large language models to zero-shot and few-shot classification of tabular data \cite{hegselmann2023tabllm}. TabPFN shows strong few-shot performance on tabular data by pretraining on a large number of synthetic datasets \cite{hollmann2023tabpfn}. LaTable demonstrates potential for zero-shot classification results by leveraging dataset and column descriptions \cite{van2024latable}. Prior works such as \cite{wu2025zeroshot} focus on zero-shot meta learning, they use the term to refer to models that perform predictions on unseen datasets with zero gradient update on the model, and with no reliance on the context similarity between this dataset and the datasets that the model was pre-trained on. However, this concept diverges from the typical zero-shot learning. In our work, zero-shot learning refers to the amount of empirical examples for prediction being zero. 

\textbf{Model-free feature importance.} Model-free feature importance methods aim to quantify the contribution of each feature without relying on a fitted predictive model, distinguishing them from model-agnostic methods that still require model evaluations. A central example is the Leave-One-Covariate-Out (LOCO) framework of \citet{lei2018distribution}, which measures a feature’s importance by assessing how predictive performance changes when that feature is removed from the dataset. However, LOCO is only able to provide global feature contributions and still requires knowledge about the base model in order to train comparable surrogate models without the feature of interest. Subsequent work has refined this idea: \citet{verdinelli2024feature} provide improved statistical guarantees and broaden the applicability of LOCO-style estimators, while \citet{williamson2023general} introduce a generalized formulation that incorporates a data-splitting strategy to yield more stable and interpretable importance estimates. Extending the LOCO intuition beyond any specific model class,~\citet{williamson2021nonparametric} propose a nonparametric variable-importance measure that uses cross-fitting and influence-function theory to deliver unbiased global estimates and confidence intervals.

\textbf{Explanation Multiplicity.} \citet{hwang2025shap} show that SHAP-based feature importance explanations are sensitive to feature representation and simple preprocessing choices, indicating that explanation methods themselves are unstable to seemingly innocuous changes in input encoding. Similarly,~\citet{slack2021counterfactual} demonstrate that counterfactual explanations are highly sensitive to input perturbations and can be manipulated, revealing that explanations may not be reliable under predictive multiplicity and can conceal model biases even when test accuracy is unchanged. More broadly, the model multiplicity phenomenon has been formalized and explored across several studies. \citet{black2022model} define and analyze model multiplicity, showing there can exist many predictively equivalent models with different decision boundaries, which has important implications for fairness, arbitrariness, and recourse. In the context of specific model classes,~\citet{xin2022exploring} develop methods to enumerate the Rashomon set of sparse decision trees, making it possible to inspect the full set of nearly equally good models rather than selecting only one. Building on this,~\citet{kobylinska2024exploration} show that it is possible to identify models with substantially different behavior by exploring the Rashomon set, which helps generate more trustworthy explanations in high-stakes domains like medical data analysis. Together, these works illustrate both the risks of relying on a single model/explanation and the opportunities afforded by explicitly considering model and explanation multiplicity when building and interpreting machine learning systems. 

\section{Feature Attribution Paradigms: true to the model vs. true to the data}
\label{app:true-to-data}

A well-known tension in Shapley-based feature attribution concerns whether explanations should reflect the behavior of the model or the structure of the data distribution~\citep{chen2020true}. The distinction arises from a fundamental design choice in computing Eq.~(\ref{eq:shapley}): how to handle absent features when evaluating coalitions.

The \emph{marginal} approach, used by default Kernel SHAP~\citep{lundberg2017unified}, treats absent features as independent of observed ones, sampling from the marginal training distribution. This is \emph{true to the model} in that it queries the model on any input regardless of plausibility, but it routinely evaluates feature combinations that never arise in practice.  \emph{Conditional} approaches~\citep{aas2021explaining} instead sample absent features from their conditional distribution given observed features, capturing inter-feature dependencies but inheriting errors from the density estimator. \emph{Causal} approaches~\citep{heskes2020causal} propagate interventions through an assumed causal graph, and \emph{on-manifold} approaches~\citep{fryeshapley} restrict evaluation to the support of the data distribution. The latter three each attempt, in different ways, to remain \emph{true to the data}.

Crucially, all four approaches share a common prerequisite: access to the target model $f^b$ to evaluate feature coalitions. ExplainerPFN occupies a categorically different position. Because it has no access to $f^b$, model-consistent attribution is not a design choice to be weighed against data-consistent alternatives---it is simply unavailable. ExplainerPFN instead learns a posterior mean attribution from the data distribution and predictions alone, targeting explanations \emph{true to the data} by necessity. In the zero-access settings ExplainerPFN is designed for, the true-to-the-model vs.\ true-to-the-data axis collapses to a single option, and ExplainerPFN is the first method to occupy it.
\section{Additional details on training data generation}\label{sec:appendix_data_generation}

This section extends Section~\ref{sec:training-data} with additional details on the synthetic data generation process used to train ExplainerPFN. In particular, we describe the procedure used to sample directed acyclic graphs (DAGs), the distributions employed to generate values for exogenous (base) nodes, and the propagation mechanisms used to compute values for downstream nodes.

\emph{DAG Sampling.} To generate structural dependencies among features, we sample directed acyclic graphs (DAGs) using the Growing Network with Redirection (GNR) procedure~\cite{krapivsky2001organization}, which produces sparse, scale-free graph topologies. In this framework, nodes are added sequentially; each new node selects an existing node uniformly at random and either attaches to it directly or, with probability $p$, sampled from either a gamma or uniform distribution, redirects the edge to that node’s parent. This process naturally induces heterogeneous in-degree distributions and a mixture of shallow and deep dependency chains, providing diverse structural patterns for pretraining. The number of nodes $N$ is sampled from a uniform distribution over a user-specified range (we chose $[2, 10]$), while the redirection probability $p$ is drawn from a Gamma distribution, allowing control over the expected sparsity of the resulting DAG. We also allow for the generation of multiple, small, disconnected subgraphs, generated as described previously, which are then connected following GNR procedure. When selecting the target node and input features, we allow for the possibility that some features may be disconnected from the target node (i.e., belong to different subgraphs). This reflects real-world scenarios where some features may be non-informative for the learning task.

\emph{Edge mappings.} Each edge in the sampled DAG is associated with a nonlinear activation used when propagating values from parent to child nodes. We draw these activation functions uniformly from a diverse set to encourage nontrivial feature interactions: identity, logarithm, sigmoid, absolute value, sine, hyperbolic tangent, rank transformation, square and power functions, SmoothReLU/Softplus~\cite{dugas2000incorporating}, step, and modulo operations.

\emph{Computational considerations.} Due to the computational overhead introduced with the training of a ML classifier, as well as computation of Shapley values for each classifier and synthetic dataset, the synthetic data generation process cannot be done online (i.e., as the model is being trained). Therefore, we generate data in parallel to the model training; we instantiated two VMs, where one is generating and saving data, while the other is sampling data from the generation process. This means ExplainerPFN can repeatedly sample the same synthetic datasets over the generation process. We attempted to mitigate this issue via the generation of large amounts of data (hundreds of thousands of datasets).

\emph{Future work.} We note that the current generator uses only a fixed family of edge mappings. As part of future work, we plan to expand this library to incorporate additional distributional transformations (e.g., Kumaraswamy distortions~\cite{kumaraswamy1980generalized}, rule-based dependencies, etc.) and more complex base node sampling mechanisms to account for between-observation dependencies, to further increase the quality of the synthetic datasets. In addition, we plan to explore other mechanisms to mitigate some of the computational overhead introduced by the data generation process.

\begin{table*}[htb]
    \centering
    \caption{Pearson correlation between true Shapley values computed based on a MLP (as the base predictor) and estimated Shapley values using ExplainerPFN (zero-shot predictions), as well as a MLP, a RF and TabPFN in different few-shot settings with varying numbers of training samples. Results are reported across a diverse set of UCI datasets.}
    \label{tab:results_corr_across_datasets_mlp}
    \vspace{0.5em}
    \footnotesize
    \resizebox{\textwidth}{!}{\begin{tabular}{ccccccccccccc}
\toprule
Method & Samples & BA & AB & AN & CC & EC & FL & GC & HE & HD & HP & TH \\
\midrule
\multirow[c]{1}{*}{EPFN} & 0 & 0.880 & 0.114 & \textbf{0.338} & 0.530 & 0.490 & 0.058 & 0.213 & 0.463 & \textbf{0.557} & \textbf{0.485} & 0.212 \\
\midrule
\multirow[c]{5}{*}{MLP} & 2 & -0.052 & -0.005 & -0.046 & 0.032 & 0.064 & 0.102 & 0.178 & -0.131 & 0.210 & -0.023 & 0.045 \\
 & 4 & 0.027 & 0.221 & 0.153 & 0.067 & 0.216 & 0.068 & 0.157 & -0.170 & 0.053 & 0.056 & 0.050 \\
 & 6 & 0.088 & 0.155 & 0.151 & 0.051 & 0.363 & 0.147 & 0.104 & -0.151 & 0.119 & 0.039 & 0.053 \\
 & 8 & 0.744 & 0.180 & 0.150 & 0.076 & 0.329 & 0.173 & 0.163 & -0.136 & 0.192 & 0.054 & 0.057 \\
 & 10 & 0.734 & 0.144 & 0.151 & 0.074 & 0.466 & 0.167 & 0.143 & -0.127 & 0.080 & 0.055 & 0.055 \\
\midrule
\multirow[c]{5}{*}{RF} & 2 & 0.207 & 0.021 & 0.030 & 0.166 & 0.195 & 0.098 & 0.055 & -0.119 & 0.172 & -0.002 & 0.053 \\
 & 4 & 0.364 & 0.092 & 0.051 & 0.312 & 0.362 & 0.250 & 0.221 & 0.158 & 0.213 & 0.039 & 0.121 \\
 & 6 & 0.437 & 0.628 & 0.172 & 0.273 & 0.734 & 0.374 & 0.311 & 0.242 & 0.259 & 0.025 & 0.342 \\
 & 8 & 0.775 & \textbf{0.687} & 0.263 & 0.453 & 0.682 & 0.416 & 0.342 & 0.320 & 0.386 & 0.046 & 0.373 \\
 & 10 & 0.830 & \underline{0.684} & \underline{0.290} & 0.561 & 0.909 & \underline{0.462} & 0.355 & 0.386 & \underline{0.456} & 0.044 & \underline{0.608} \\
\midrule
\multirow[c]{5}{*}{TabPFN} & 2 & 0.073 & 0.022 & 0.033 & 0.061 & 0.074 & 0.054 & 0.049 & -0.145 & 0.008 & -0.013 & 0.050 \\
 & 4 & 0.714 & 0.149 & 0.020 & \underline{0.632} & 0.808 & 0.396 & 0.282 & 0.163 & 0.328 & 0.054 & 0.143 \\
 & 6 & 0.837 & 0.599 & 0.064 & 0.439 & 0.902 & 0.461 & 0.258 & 0.392 & 0.358 & 0.078 & 0.329 \\
 & 8 & \underline{0.903} & 0.607 & 0.156 & 0.515 & \underline{0.912} & 0.450 & \underline{0.366} & \underline{0.513} & 0.428 & \underline{0.101} & 0.346 \\
 & 10 & \textbf{0.956} & 0.643 & 0.167 & \textbf{0.790} & \textbf{0.925} & \textbf{0.500} & \textbf{0.445} & \textbf{0.538} & 0.214 & 0.088 & \textbf{0.693} \\
\bottomrule
\end{tabular}}
\end{table*}

\begin{table*}[htb]
    \centering
    \caption{Spearman rank correlation between true Shapley values computed based on a MLP (as the base predictor) and estimated Shapley values using ExplainerPFN (zero-shot predictions), as well as a MLP, a RF and TabPFN in different few-shot settings with varying numbers of training samples.}
    \label{tab:results_spearman_across_datasets_mlp}
    \vspace{0.5em}
    \footnotesize
    \resizebox{\textwidth}{!}{\begin{tabular}{ccccccccccccc}
\toprule
Method & Samples & BA & AB & AN & CC & EC & FL & GC & HE & HD & HP & TH \\
\midrule
\multirow[c]{1}{*}{EPFN} & 0 & 0.920 & 0.185 & 0.306 & 0.522 & 0.534 & 0.244 & 0.211 & \underline{0.405} & \underline{0.466} & -0.182 & 0.134 \\
\midrule
\multirow[c]{5}{*}{MLP} & 2 & 0.059 & 0.089 & 0.002 & 0.030 & -0.019 & 0.212 & 0.179 & -0.162 & 0.199 & -0.017 & 0.005 \\
 & 4 & 0.178 & 0.231 & 0.021 & 0.093 & 0.051 & 0.111 & 0.093 & -0.189 & 0.245 & 0.196 & 0.047 \\
 & 6 & 0.324 & 0.053 & -0.002 & 0.095 & 0.119 & 0.085 & 0.030 & -0.176 & 0.214 & 0.160 & 0.057 \\
 & 8 & 0.785 & 0.192 & 0.041 & 0.126 & 0.113 & 0.143 & 0.087 & -0.146 & 0.188 & 0.179 & 0.070 \\
 & 10 & 0.739 & 0.122 & 0.057 & 0.100 & 0.228 & 0.170 & 0.084 & -0.159 & 0.120 & 0.176 & 0.006 \\
\midrule
\multirow[c]{5}{*}{RF} & 2 & 0.298 & 0.037 & 0.300 & 0.288 & 0.122 & 0.047 & 0.122 & -0.189 & 0.194 & 0.067 & 0.042 \\
 & 4 & 0.387 & 0.097 & 0.359 & 0.227 & 0.223 & 0.224 & 0.186 & 0.023 & 0.219 & 0.293 & 0.132 \\
 & 6 & 0.499 & 0.614 & 0.481 & 0.275 & 0.479 & 0.352 & \underline{0.351} & 0.121 & 0.275 & 0.211 & 0.293 \\
 & 8 & 0.808 & 0.689 & \underline{0.550} & 0.521 & 0.508 & 0.515 & 0.303 & 0.251 & 0.392 & 0.257 & 0.396 \\
 & 10 & 0.868 & \underline{0.690} & \textbf{0.571} & \underline{0.599} & \textbf{0.696} & \underline{0.607} & \textbf{0.353} & 0.319 & \textbf{0.475} & 0.286 & 0.577 \\
\midrule
\multirow[c]{5}{*}{TabPFN} & 2 & 0.104 & 0.030 & 0.319 & 0.168 & 0.097 & 0.014 & 0.110 & -0.182 & -0.011 & 0.092 & -0.008 \\
 & 4 & 0.673 & 0.237 & 0.336 & 0.583 & 0.475 & 0.452 & 0.254 & 0.015 & 0.286 & \textbf{0.376} & 0.128 \\
 & 6 & 0.801 & 0.652 & 0.375 & 0.504 & 0.685 & 0.478 & 0.337 & 0.286 & 0.333 & 0.316 & 0.411 \\
 & 8 & \underline{0.926} & 0.681 & 0.435 & 0.557 & \underline{0.691} & 0.586 & 0.310 & 0.374 & 0.411 & 0.308 & \underline{0.619} \\
 & 10 & \textbf{0.960} & \textbf{0.716} & 0.440 & \textbf{0.805} & 0.682 & \textbf{0.672} & 0.348 & \textbf{0.446} & 0.456 & \underline{0.330} & \textbf{0.731} \\
\bottomrule
\end{tabular}}
\end{table*}

\begin{table*}[htb]
    \centering
    \caption{Spearman rank correlation between true Shapley values computed based on a RFR (as the base predictor) and estimated Shapley values using ExplainerPFN (zero-shot predictions), as well as a MLP, a RF and TabPFN in different few-shot settings with varying numbers of training samples.}
    \label{tab:results_spearman_across_datasets_rfr}
    \vspace{0.5em}
    \footnotesize
    \resizebox{\textwidth}{!}{\begin{tabular}{ccccccccccccc}
\toprule
Method & Samples & BA & AB & AN & CC & EC & FL & GC & HE & HD & HP & TH \\
\midrule
\multirow[c]{1}{*}{EPFN} & 0 & \textbf{0.923} & \textbf{0.373} & 0.298 & \textbf{0.518} & 0.373 & 0.235 & 0.115 & \textbf{0.573} & \textbf{0.466} & 0.434 & \textbf{0.672} \\
\midrule
\multirow[c]{5}{*}{MLP} & 2 & -0.002 & 0.144 & -0.047 & -0.020 & 0.112 & 0.251 & 0.118 & -0.065 & 0.240 & -0.112 & 0.058 \\
 & 4 & 0.105 & 0.176 & 0.051 & 0.099 & 0.163 & 0.127 & 0.085 & -0.140 & 0.256 & -0.107 & 0.087 \\
 & 6 & 0.236 & -0.105 & -0.023 & 0.092 & 0.184 & 0.077 & 0.033 & -0.143 & 0.229 & -0.070 & 0.089 \\
 & 8 & 0.687 & -0.019 & 0.107 & 0.109 & 0.173 & 0.122 & 0.064 & -0.093 & 0.200 & -0.039 & 0.113 \\
 & 10 & 0.632 & -0.117 & 0.119 & 0.078 & 0.212 & 0.052 & 0.076 & -0.076 & 0.140 & -0.024 & 0.066 \\
\midrule
\multirow[c]{5}{*}{RF} & 2 & 0.326 & 0.036 & 0.206 & 0.214 & -0.018 & -0.014 & 0.105 & -0.072 & 0.206 & \textbf{0.472} & -0.018 \\
 & 4 & 0.468 & 0.108 & 0.429 & 0.337 & 0.325 & 0.267 & 0.286 & 0.163 & 0.270 & \underline{0.462} & 0.017 \\
 & 6 & 0.522 & \underline{0.361} & 0.535 & 0.356 & 0.517 & 0.451 & \textbf{0.392} & 0.273 & 0.356 & 0.298 & 0.175 \\
 & 8 & 0.768 & 0.322 & \textbf{0.578} & 0.365 & 0.614 & 0.519 & 0.341 & 0.223 & 0.389 & 0.445 & 0.332 \\
 & 10 & \underline{0.911} & 0.330 & \underline{0.557} & 0.411 & \textbf{0.805} & \textbf{0.594} & \underline{0.377} & 0.341 & \underline{0.449} & 0.454 & 0.427 \\
\midrule
\multirow[c]{5}{*}{TabPFN} & 2 & 0.054 & 0.059 & 0.035 & 0.199 & -0.045 & 0.009 & 0.153 & -0.127 & -0.062 & 0.081 & -0.009 \\
 & 4 & 0.783 & 0.082 & 0.460 & \underline{0.518} & 0.460 & 0.392 & 0.220 & 0.283 & 0.277 & 0.205 & 0.144 \\
 & 6 & 0.697 & 0.320 & 0.506 & 0.329 & 0.621 & 0.460 & 0.250 & 0.294 & 0.360 & 0.193 & 0.437 \\
 & 8 & 0.754 & 0.312 & 0.542 & 0.341 & 0.747 & 0.447 & 0.357 & 0.326 & 0.412 & 0.198 & 0.627 \\
 & 10 & 0.896 & 0.314 & 0.557 & 0.459 & \underline{0.788} & \underline{0.555} & 0.302 & \underline{0.392} & 0.427 & 0.268 & \underline{0.637} \\
\bottomrule
\end{tabular}}
\end{table*}

\section{Additional information on Datasets used}\label{sec:appendix_datasets}

This section provides supplementary details on the datasets used in our experiments, including data sources, preprocessing steps, and task-specific adjustments. Throughout all experiments and preprocessing steps over all the stages of development of this research, we used a fixed random seed (42).

\begin{table} 
    \centering
    \caption{Summary of datasets used in experiments.}
    \label{tab:dataset_summary}
    \vspace{0.5em}
    \footnotesize
    \begin{tabular}{ccccc}
\toprule
Dataset name & Obs. & Feat. & Maj./Min. & IR \\
\midrule
Echocardiogram (EC) & 61 & 8 & 44/17 & 2.59 \\
Hepatitis (HP) & 129 & 4 & 105/24 & 4.38 \\
Flags (FL) & 194 & 10 & 134/60 & 2.23 \\
Heart (HE) & 270 & 6 & 150/120 & 1.25 \\
Heart Disease (HD) & 740 & 5 & 383/357 & 1.07 \\
Annealing (AN) & 798 & 6 & 608/190 & 3.20 \\
German Credit (GC) & 1000 & 7 & 700/300 & 2.33 \\
Banknote Auth. (BA) & 1372 & 4 & 762/610 & 1.25 \\
Contraceptive (CC) & 1473 & 5 & 844/629 & 1.34 \\
Abalone (AB) & 4177 & 7 & 3488/689 & 5.06 \\
Thyroid (TH) & 5789 & 6 & 4168/1621 & 2.57 \\
\bottomrule
\end{tabular}

\end{table}

\subsection{Tabular Benchmark Datasets}

We evaluate ExplainerPFN on a collection of tabular binary classification tasks, focusing on continuous features. For datasets with originally multiclass classification tasks, we convert the task to binary classification using a majority-vs-rest scheme. To maintain consistency with TabPFN-style problem settings and ExplainerPFN's training scheme, we retain only datasets satisfying $2 < m < 15$ and $n < 20{,}000$.

During evaluation, modeling pipelines include a standard scaler before fitting the downstream classifier or regressor. Shapley values are computed on the scaled feature representations, since the SHAP framework does not support explanations over the raw input data under certain scenarios.

\subsection{ACS Public Coverage Case Study}

For the ACS case study, we use the 2018 American Community Survey (ACS) data for Alabama, accessed via the folktables interface~\cite{ding2021retiring}. We restrict the feature set to AGEP, SCHL, PINCP, and SEX to facilitate analysis while retaining sensitive attributes. The dataset is split into a training and test set, both with 2000 samples. We train a pipeline consisting of a standard scaler, followed by a tuned MLP. SHAP values are computed on the fitted model, while ExplainerPFN uses the concatenated train and test feature matrix, along with the model's scores.

\subsection{DAG Recovery Datasets}

For the structure-recovery experiments, we use synthetic data generated by the DAG process described in Section~\ref{sec:training-data} and Appendix~\ref{sec:appendix_data_generation}. After generation, the designated target column is removed before computing feature importance estimates. Explanations are computed independently for each feature. We construct directed edges by thresholding these magnitudes at a fixed percentile level. This procedure yields an influence graph that can be directly compared to the ground-truth DAG.

\section{Additional Details on Improving Attribution Consistency via Shapley Properties}
\label{app:post-processing}

Shapley values are characterized by axioms including symmetry (features that play equivalent roles should receive equivalent attributions) and efficiency (the total attribution equals the model output, up to a baseline)~\cite{shapley1953value}, which impose structural constraints on valid feature attributions. Although ExplainerPFN is trained to predict Shapley values directly, its raw outputs need not satisfy these properties exactly. We use the Shapley axioms as calibration constraints, applying a post-processing step to improve between-feature comparability and explanation fidelity.  Each correction below enforces a consequence of the Shapley axioms that can be imposed using only the predicted attributions and model outputs, without access to the underlying model.

\textbf{Mean-zero attributions (efficiency in expectation).} Given $X$, $\hat{Y}$, and the base value $v = E[\hat{Y}] = E[f^b(X)]$, we find that $\hat{y}_i = v + \sum_{j=1}^m \phi_i^j$. Using empirical expectations over the dataset as estimators of population expectations gives $E[\hat{Y}] - v = E[\sum_{j=1}^m \phi^j] = 0$. Thus, the feature attributions collectively have zero mean across the dataset, $\sum_{i=1}^n \sum_{j=1}^m \phi_i^j = 0$. This allows us to apply a simple re-centering step to the predicted attributions, $\tilde{\phi}_i^j = \hat{\phi}_i^j - \frac{1}{n.m} \sum_{i=1}^n \sum_{j=1}^m \hat{\phi}_i^j$.

\textbf{Variance decomposition (approximate symmetry via between-feature scaling).} SHAP commonly assumes that input features are independent when computing feature attributions. In practice, if we also treat feature attributions as approximately independent (which is consistent with the assumptions in Linear-SHAP~\cite{lundberg2017unified}), \emph{i.e.}, $Cov(\phi^a, \phi^b) = 0 \forall \{a, b\} \in M, a \neq b$, then the variance of the model's predictions can be decomposed as $Var(\hat{Y}) = \sum_{j=1}^m Var(\phi^j)$. Equivalently, we can consider the expected variance of each feature attribution as $\sigma^2 = \frac{Var(\hat{Y})}{m}$, and standard deviation $\sigma = \frac{Std(\hat{Y})}{\sqrt{m}}$. We rescale the predicted attributions to match the expected per-feature variance,
\[
\tilde{\phi}_i^j = \hat{\phi}_i^j \cdot \frac{Std(\hat{Y}) / \sqrt{m}}{Std(\hat{\Phi})},
\]
where $Std(\hat{\Phi})$ denotes the standard deviation of all predicted attributions in the dataset.

\textbf{Instance-level efficiency.} To ensure $\hat{y}_i = v + \sum_{j=1}^m \tilde{\phi}^j_i$, we introduce an error term $\epsilon_i$, computed as:
\begin{equation}
  \hat{y}_i = v + \sum_{j=1}^m[\hat{\phi}_i^j + \epsilon_i] \iff \epsilon_i = \frac{\hat{y}_i - v - \sum_{j=1}^m \hat{\phi}_i^j}{m},
\end{equation}
and added to $\hat{\phi}_i^j$ if no statistical corrections are applied.

\textbf{Post-processed attributions.} Finally, combining all three steps yields the adjusted predicted attributions. We define the partially corrected attributions as:
\begin{equation}
  \label{eq:post_processed_attributions}
  \phi_i^{j, \text{partial}} = \left( \hat{\phi}_i^j - \frac{1}{n.m} \sum_{i=1}^n \sum_{j=1}^m \hat{\phi}_i^j \right) \cdot \frac{Std(\hat{Y}) / \sqrt{m}}{Std(\hat{\Phi})}.
\end{equation}
Over which we apply $\epsilon_i$, \emph{after} the statistical corrections:
\begin{equation}
  \tilde{\phi}_i^{j} = \phi_i^{j, \text{partial}} + \frac{\hat{y}_i - v - \sum_{j=1}^m \phi_i^{j, \text{partial}}}{m}.
\end{equation}

\section{Additional experimental results}\label{sec:appendix_experimental_results}

This section extends Section~\ref{sec:experiments} with additional experimental results and analyses. Following the recommendations in~\cite{hwang2026explanation}, our analyses address 4 perspectives regarding explanation evaluations:

\emph{Disentangling Sources of Multiplicity Evaluation}. Evaluations should distinguish between variations caused by model training or selection and the stochasticity inherent in the explanation pipeline itself. We addressed this concern by fixing random seeds across all model training and selection pipelines, ensuring that any observed variability in explanations arises solely from the methods under evaluation. In addition, we include an additional analysis of zero-shot results using a different base model (MLP) to assess the robustness of ExplainerPFN across varying model architectures in Appendix~\ref{sec:appendix_zero_shot_mlp}.

\emph{Magnitude-based metrics}. $\ell_2$ distance-based metrics are frequently used in the literature. We use the Pearson correlation coefficient since it is monotonically related to $\ell_2$ distance when the vectors being compared have the same mean and variance, which we ensure via our post-processing correction pipeline. These results are presented in Section~\ref{sec:experiments} and Appendix~\ref{sec:appendix_zero_shot_mlp}.

\emph{Rank-based metrics}. To better align with how explanations are used in practice, we also evaluate explanations using rank-based metrics. Specifically, we compute the top-$k$ Jaccard similarity to measure agreement in the set of features identified as the most important (positively or negatively). We include these results in Appendix~\ref{sec:appendix_rank_based_evaluation}.

\emph{Explanation evaluation as an interpretive task}. Evaluating explanations as an interpretive task rather than a purely technical or descriptive one is essential to ensure that explanations align with real-world needs. We include a case study in Section~\ref{sec:case_study} that demonstrates how ExplainerPFN compares to SHAP in a practical setting.

\begin{table*}[htb]
    \centering
    \caption{Jaccard Top-K similarity between true Shapley values computed based on a RF (as the base predictor) and estimated Shapley values using ExplainerPFN (zero-shot predictions), as well as a MLP, a RF and TabPFN in different few-shot settings with varying numbers of training samples.}
    \label{tab:results_jaccard_across_datasets_rf}
    \vspace{0.5em}
    \footnotesize
    \resizebox{\textwidth}{!}{\begin{tabular}{ccccccccccccc}
\toprule
Method & Samples & BA & AB & AN & CC & EC & FL & GC & HE & HD & HP & TH \\
\midrule
  EPFN & 0 & \underline{0.641} & 0.102 & \underline{0.588} & 0.326 & 0.298 & 0.114 & 0.118 & \textbf{0.580} & \textbf{0.559} & 0.282 & 0.272 \\
\midrule
\multirow[c]{5}{*}{MLP} & 2 & 0.257 & 0.205 & 0.329 & 0.256 & 0.263 & 0.310 & 0.226 & 0.214 & 0.104 & 0.128 & 0.563 \\
 & 4 & 0.515 & 0.151 & 0.318 & 0.213 & 0.404 & 0.256 & 0.131 & 0.230 & 0.054 & 0.128 & 0.320 \\
 & 6 & 0.631 & 0.142 & 0.294 & 0.319 & 0.316 & 0.239 & 0.208 & 0.272 & 0.225 & 0.179 & 0.372 \\
 & 8 & 0.437 & 0.235 & 0.435 & 0.296 & 0.351 & 0.249 & 0.249 & 0.235 & 0.099 & 0.179 & 0.324 \\
 & 10 & 0.333 & 0.197 & 0.383 & 0.276 & 0.456 & 0.263 & 0.239 & 0.202 & 0.122 & 0.154 & 0.367 \\
\midrule
  \multirow[c]{5}{*}{RF} & 2 & 0.138 & 0.191 & 0.356 & 0.421 & 0.404 & 0.466 & 0.299 & 0.370 & 0.140 & 0.256 & \textbf{0.812} \\
  & 4 & 0.439 & \textbf{0.367} & 0.446 & \textbf{0.455} & 0.649 & 0.414 & \textbf{0.400} & 0.317 & 0.185 & \textbf{0.308} & 0.729 \\
  & 6 & 0.510 & 0.266 & 0.496 & 0.378 & 0.579 & \underline{0.488} & 0.372 & 0.358 & 0.189 & 0.231 & 0.633 \\
 & 8 & 0.595 & 0.314 & 0.562 & 0.400 & 0.579 & 0.437 & 0.384 & 0.329 & 0.176 & 0.231 & 0.680 \\
  & 10 & 0.580 & 0.325 & \textbf{0.590} & 0.382 & \underline{0.702} & \textbf{0.520} & 0.366 & 0.362 & 0.216 & 0.231 & 0.665 \\
\midrule
  \multirow[c]{5}{*}{TabPFN} & 2 & 0.170 & 0.191 & 0.300 & 0.387 & 0.404 & 0.341 & \underline{0.397} & 0.403 & 0.225 & 0.205 & \underline{0.811} \\
  & 4 & 0.311 & \underline{0.364} & 0.454 & \underline{0.437} & 0.579 & 0.454 & 0.382 & 0.374 & 0.194 & \textbf{0.308} & 0.742 \\
 & 6 & 0.578 & 0.236 & 0.469 & 0.351 & 0.579 & 0.478 & 0.332 & 0.416 & 0.167 & 0.282 & 0.630 \\
  & 8 & 0.614 & 0.306 & 0.508 & 0.373 & 0.614 & 0.459 & 0.393 & \underline{0.440} & 0.212 & 0.256 & 0.770 \\
  & 10 & \textbf{0.750} & 0.304 & 0.521 & 0.378 & \textbf{0.825} & 0.458 & 0.323 & 0.436 & \underline{0.252} & 0.256 & 0.759 \\
\bottomrule
\end{tabular}
}
\end{table*}

\begin{table*}[htb]
    \centering
    \caption{Jaccard Top-K similarity between true Shapley values computed based on a MLP (as the base predictor) and estimated Shapley values using ExplainerPFN (zero-shot predictions), as well as a MLP, a RF and TabPFN in different few-shot settings with varying numbers of training samples.}
    \label{tab:results_jaccard_across_datasets_mlp}
    \vspace{0.5em}
    \footnotesize
    \resizebox{\textwidth}{!}{\begin{tabular}{ccccccccccccc}
\toprule
Method & Samples & BA & AB & AN & CC & EC & FL & GC & HE & HD & HP & TH \\
\midrule
  EPFN & 0 & 0.738 & 0.461 & 0.371 & 0.428 & 0.474 & 0.269 & 0.170 & \textbf{0.617} & \textbf{0.505} & 0.077 & 0.453 \\
\midrule
  \multirow[c]{5}{*}{MLP} & 2 & 0.381 & 0.270 & 0.258 & 0.251 & 0.614 & 0.247 & 0.216 & 0.222 & 0.099 & \textbf{0.256} & 0.423 \\
  & 4 & 0.318 & 0.225 & 0.304 & 0.294 & 0.526 & 0.242 & 0.232 & 0.206 & 0.212 & \underline{0.231} & 0.277 \\
 & 6 & 0.515 & 0.278 & 0.349 & 0.348 & 0.544 & 0.249 & 0.191 & 0.210 & 0.171 & 0.179 & 0.311 \\
 & 8 & 0.709 & 0.340 & 0.343 & 0.337 & 0.526 & 0.239 & 0.223 & 0.210 & 0.095 & 0.205 & 0.307 \\
 & 10 & 0.575 & 0.305 & 0.346 & 0.324 & 0.421 & 0.263 & 0.259 & 0.173 & 0.126 & 0.205 & 0.331 \\
\midrule
  \multirow[c]{5}{*}{RF} & 2 & 0.175 & 0.239 & \textbf{0.433} & 0.606 & 0.404 & 0.341 & \textbf{0.337} & \underline{0.337} & 0.122 & 0.179 & \textbf{0.612} \\
 & 4 & 0.345 & 0.548 & 0.399 & 0.638 & 0.544 & 0.354 & 0.307 & 0.222 & 0.144 & 0.179 & 0.581 \\
 & 6 & 0.367 & 0.506 & 0.338 & 0.631 & 0.649 & 0.381 & 0.321 & 0.284 & 0.185 & 0.179 & 0.558 \\
 & 8 & 0.600 & 0.533 & 0.353 & 0.629 & 0.632 & 0.368 & 0.299 & 0.243 & 0.198 & 0.179 & 0.586 \\
 & 10 & 0.602 & 0.537 & 0.365 & 0.570 & 0.719 & 0.380 & 0.258 & 0.267 & 0.212 & 0.154 & 0.587 \\
\midrule
  \multirow[c]{5}{*}{TabPFN} & 2 & 0.194 & 0.239 & 0.424 & 0.570 & 0.263 & 0.353 & \textbf{0.337} & 0.329 & 0.108 & 0.179 & \underline{0.611} \\
  & 4 & 0.706 & 0.401 & \underline{0.428} & 0.597 & 0.509 & 0.285 & 0.314 & 0.214 & 0.135 & 0.077 & 0.585 \\
 & 6 & 0.731 & 0.407 & 0.412 & 0.602 & 0.719 & 0.398 & 0.300 & 0.292 & 0.207 & 0.077 & 0.563 \\
  & 8 & \underline{0.811} & \underline{0.478} & 0.396 & \textbf{0.658} & \textbf{0.754} & \underline{0.412} & 0.288 & 0.325 & 0.239 & 0.103 & 0.607 \\
  & 10 & \textbf{0.920} & \textbf{0.497} & 0.397 & \underline{0.656} & \textbf{0.754} & \textbf{0.466} & 0.234 & 0.317 & \underline{0.243} & 0.179 & 0.610 \\
\bottomrule
\end{tabular}
}
\end{table*}

\subsection{Zero-shot results with MLP as base model}\label{sec:appendix_zero_shot_mlp}

Table~\ref{tab:results_corr_across_datasets_mlp} presents analogous results to Section~\ref{sec:exp_zero_shot}, but using a MLP as $f^b$. ExplainerPFN achieves competitive performance across most datasets, with particularly strong performance on 4 datasets. Notably, on three datasets (Annealing, Heart Disease, and Hepatitis), ExplainerPFN outperforms all few-shot methods regardless of the number of training samples. However, performance varies substantially across datasets, with weaker results on datasets such as Abalone and Flags, suggesting that certain data characteristics may pose challenges for zero-shot estimation. 

\subsection{Rank-based evaluation}\label{sec:appendix_rank_based_evaluation}

We evaluate rank-based explanation quality using the Jaccard top-K similarity metric, which measures the agreement between the sets of the top-K most important features (based on absolute feature importance values) from the estimated and reference Shapley values. Following the definition in~\cite{hwang2026explanation}, given two rankings $\pi_i$ and $\pi_j$ induced by two feature importance vectors $\phi_i$ and $\phi_j$, we define the Jaccard Top-K similarity using the top-k features, $\pi_i^{k}$ and $\pi_j^{k}$, as follows:
\begin{equation}
    JaccardSimilarity(\pi_i, \pi_j, k) =
    \frac{|\pi_i^{k} \cap \pi_j^{k}|}{|\pi_i^{k} \cup \pi_j^{k}|},
\end{equation}

Where a value of 1 indicates perfect agreement between the two sets of top-K features, while a value of 0 indicates no overlap. In our experiments, we set K to be one-third of the total number of features, i.e., $K = \max(1, \lfloor\frac{1}{3} |\phi|\rfloor)$.

Overall, the Jaccard Top-K results presented in Table~\ref{tab:results_jaccard_across_datasets_rf}, using RF as the base model, and Table~\ref{tab:results_jaccard_across_datasets_mlp}, using an MLP as the base model, align with the findings discussed in the case study analysis in Section~\ref{sec:case_study}: ExplainerPFN is able to identify key features in a zero-shot manner in several cases, but the between-feature scaling and offset limitations previously discussed become evident.

\begin{figure}[t!]
    \centering
    \begin{subfigure}{0.20\textwidth}
        \centering
        \includegraphics[width=\linewidth]{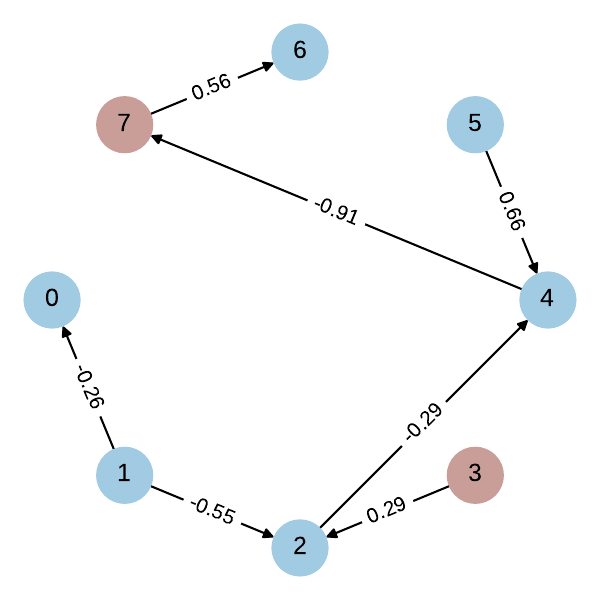}
        \caption{}
        \label{fig:dag_ground_truth}
    \end{subfigure}
    \hfill
    \begin{subfigure}{0.20\textwidth}
        \centering
        \includegraphics[width=\linewidth]{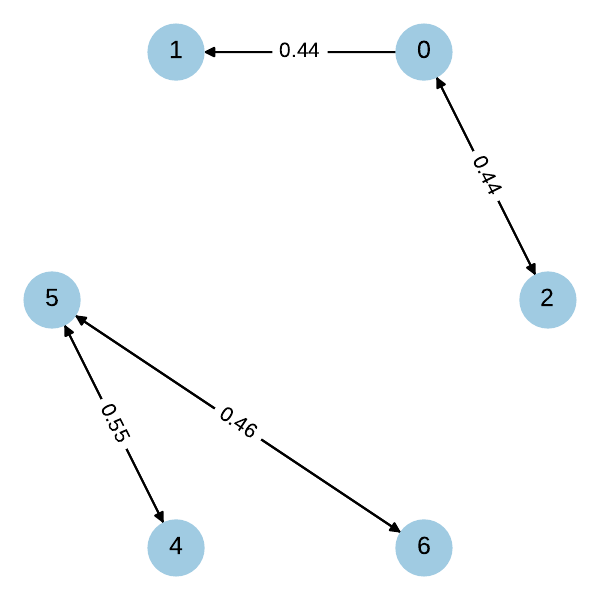}
        \caption{}
        \label{fig:dag_predicted}
    \end{subfigure}
    \hfill
    \begin{subfigure}{0.25\textwidth}
        \centering
        \includegraphics[width=\linewidth]{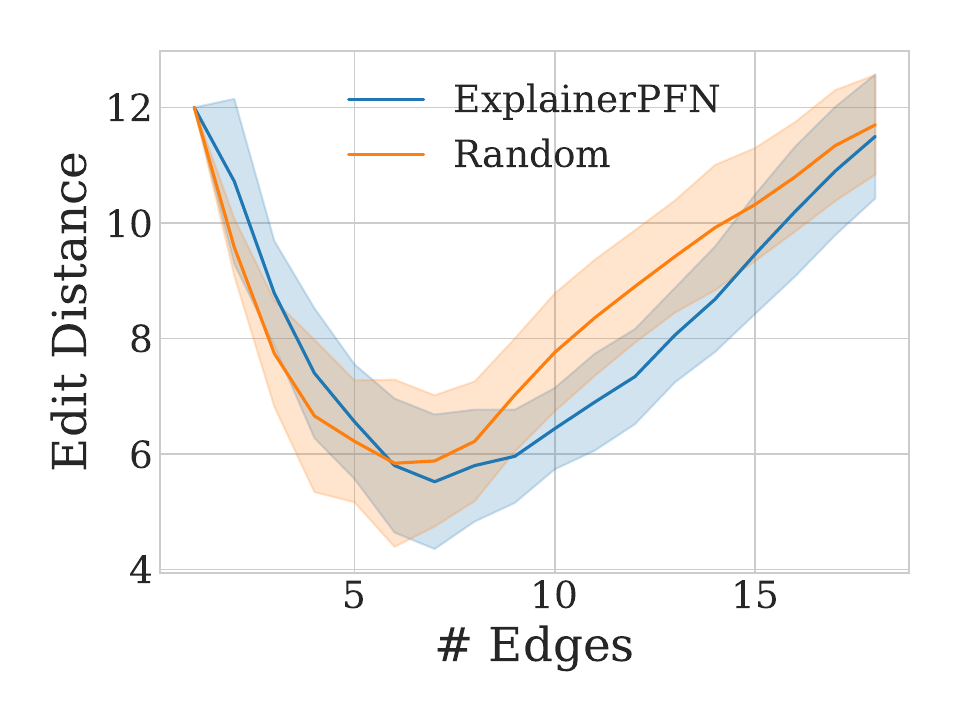}
        \caption{}
        \label{fig:dag_distance}
    \end{subfigure}
    \caption{DAG reconstruction analyses. (a) Example of a ground truth DAG. The nodes/features marked in red are removed after generating synthetic data, but before feeding the resulting data to TabPFN. (b) Estimated DAG using ExplainerPFN with the highest-scoring 7 edges, based on feature contributions estimated by ExplainerPFN. (c) Graph Edit Distance between the true and reconstructed DAGs as a function of the number of edges kept when reconstructing the DAG using ExplainerPFN over 50 synthetic datasets generated with randomly sampled DAGs.}
    \label{fig:dag_reconstruction}
\end{figure}

\subsection{Computational Efficiency}

Beyond accuracy, computational efficiency is a critical consideration for practical deployment of feature attribution methods. Table~\ref{tab:results_time_across_datasets} shows a significant gap between SHAP and the learned estimators, with the magnitude of this gap strongly dependent on the base predictor's complexity. When the base predictor is a small MLP, SHAP achieves the second fastest computation time (0.234s), outperformed by the MLP surrogate (0.106s). For small neural networks, the cost of forward passes during SHAP's sampling procedure remains low. However, as the base predictor complexity increases to a RF with 500 estimators, SHAP's computation time increases to $\approx$8.5 seconds, a 36$\times$ slowdown. In contrast, ExplainerPFN's inference time depends only on the input data. As a result, its inference time remains nearly constant across both base predictors, representing $>$10$\times$ speedup over SHAP when explaining RF predictions.

\begin{table}
    \centering
    \caption{Average computation time (in seconds) for estimating 1000 feature contributions using different methods across all tested datasets. Inference times for TabPFN, MLP and RF are reported for 10-shot settings and include the cost of computing SHAP values for the supervision samples.}
    \label{tab:results_time_across_datasets}
    \vspace{0.5em}
    \footnotesize
    \begin{tabular}{ccc}
\toprule
   & \multicolumn{2}{c}{Base Model}  \\
  Method & MLP & RF \\
  \midrule
  SHAP & \underline{1.629} & 94.872 \\
  ExplainerPFN & 2.747 & \underline{2.338} \\
  TabPFN & 2.892 & 4.717 \\
  MLP & \textbf{0.254} & \textbf{1.948} \\
  RF & 2.340 & 4.062 \\
\bottomrule
\end{tabular}
\end{table}

These results reveal an accuracy-efficiency trade-off. ExplainerPFN achieves zero-shot performance frequently superior to 4-6 shot surrogate explainers (see Tables~\ref{tab:results_corr_across_datasets_rfr} and~\ref{tab:results_corr_across_datasets_mlp}), while maintaining inference speed comparable to or faster than methods that rely on supervised SHAP examples. While SHAP provides high-quality estimations when the base model has reduced inference time and accessible, ExplainerPFN offers a practical alternative for scenarios where computational resources are limited, model access is restricted, or real-time explanations are required.

\subsection{Structural Causal Model Recovery}\label{sec:exp_dag_structure}

We investigate whether ExplainerPFN can recover structural relationships among features in synthetic datasets generated from known DAGs. To do this, we compute graph edit distance (GED) to evaluate ExplainerPFN’s ability to reconstruct the true DAG structure of the underlying data-generating process over synthetic datasets.

Although ExplainerPFN is not trained to perform causal discovery, this experiment highlights the potential of zero-shot feature-importance estimation for uncovering dependencies between variables. This connection is motivated by the meta-training setup in Section~\ref{sec:training-data}, where each synthetic task is sampled from a SCM. Because ExplainerPFN learns to map feature interactions to Shapley value patterns without access to an explicit predictive model $f^b$, we hypothesize that its output may indirectly encode the underlying structural relationships.

To assess this, we estimate feature contributions for each feature $x^j$ using a zero-shot attribution pass, $\Phi^j = F_{zs}(D \setminus \{x^j\}, x^j)$, followed by the post-processing step in Equation~\ref{eq:post_processed_attributions}. We then aggregate the feature attributions of feature $k$ as a predictor of feature $j$ with $\phi^j_{\text{agg}} = \{ E[|\phi^k | ] \mid k \in M \setminus \{j\}\}$, where $E[|\phi^k|]$. This produces a weighted directed graph whose nodes represent features and edges represent the strength of feature contributions. 

The results shown in Figure~\ref{fig:dag_reconstruction} illustrate the potential of ExplainerPFN to reconstruct the underlying DAG structure of synthetic datasets. Figure~\ref{fig:dag_ground_truth} presents a DAG, while Figure~\ref{fig:dag_predicted} shows its estimated reconstruction using ExplainerPFN after keeping the highest-scoring 7 edges. We observe that ExplainerPFN finds two strongly connected sets of features $\{4, 5, 6\}$ and $\{0, 1, 2\}$, but fails to connect the two sets. Figure~\ref{fig:dag_distance} presents the graph edit distance (GED) between the true and reconstructed DAGs, as well as DAGs constructed with uniformly sampled feature importances (random), as a function of the number of edges kept when reconstructing the DAG over 50 synthetic datasets. These datasets were generated with randomly sampled DAGs, showing that ExplainerPFN is able to provide non-random DAG reconstructions, but with mild success.

\section{Ablations}\label{sec:appendix_ablation}

\begin{figure*}[htb]
    \centering
    \includegraphics[width=\linewidth]{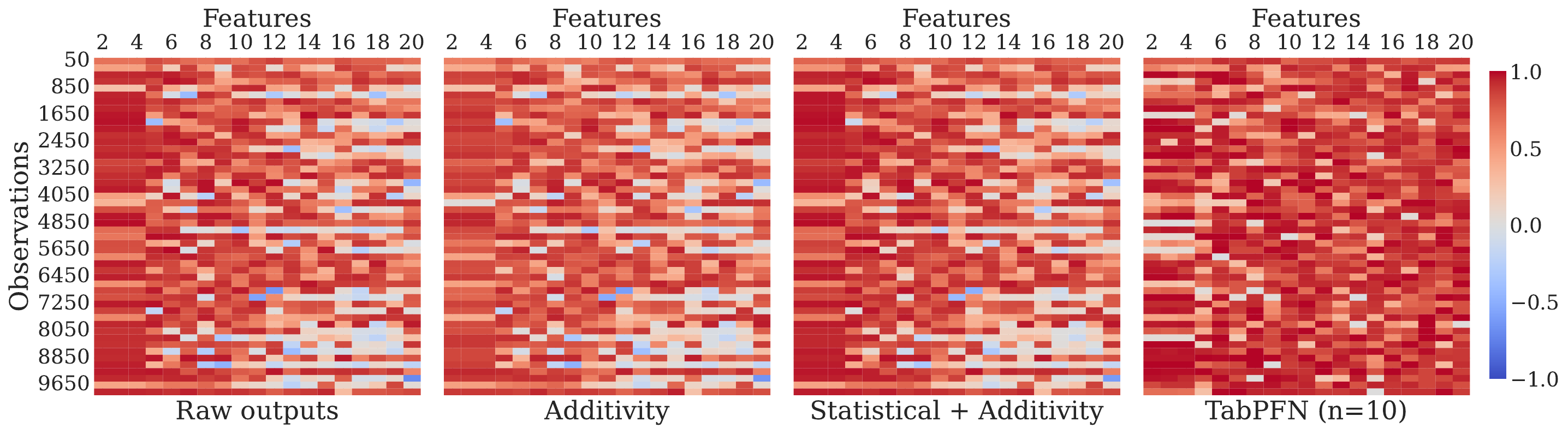}
    \caption{Pearson correlations between true and estimated Shapley values across different synthetic datasets for TabPFN with 10 training samples and ExplainerPFN outputs, including the raw outputs and correction steps.}
    \label{fig:ablation_heatmap}
\end{figure*}

The analysis in this section were conducted on synthetic datasets generated using the procedure described in Section~\ref{sec:training-data} and Appendix~\ref{sec:appendix_data_generation}. We generated datasets with varying numbers of features $[2, 20]$ and observations $[50, 10000]$ at step sizes of 1 and 200, respectively. For each combination of features and observations, we trained a base model and computed the corresponding SHAP explanations, which we used to evaluate ExplainerPFN's zero-shot estimation quality over a total of 950 datasets, corresponding to combination between 19 feature sizes and 50 observation sizes. Our ablation studies highlight several insights and directions for future research.

First, the results in Figure~\ref{fig:ablation_combined} indicate that model performance is more sensitive to the number of features than to the number of observations. Although we observe substantial variability in performance across different dataset sizes (in terms of observation count), Figure~\ref{fig:ablation_heatmap} suggests that this variability tends to increase as the number of observations grows. Overall, feature count appears to be the dominant factor influencing zero-shot estimation quality.

Second, ExplainerPFN achieves consistently high estimation quality on synthetic datasets, in contrast to the more variable performance observed on real-world data. This gap suggests that our synthetic data generator does not yet capture certain structural or statistical complexities present in real tabular distributions. Closing this gap represents an important area for improving future iterations of the synthetic data generation process. Nevertheless, the strong performance on synthetic datasets shows that this approach has potential to scale and effectively produce high quality explanations.

Third, we find that the different parts of the post-processing correction impact model performance in different ways.  The mean-zero and variance decomposition steps are not displayed standalone since they do not affect the Pearson correlation metric on their own. Instead, they improve estimation quality indirectly by aligning the scale and offset of the predicted feature attributions with the expected distribution of Shapley-based explanations, which improves the effectiveness of the additivity correction. As a result, the full correction pipeline consistently outperforms both the raw model outputs (with or without statistical normalization) and the additivity-only variant (\emph{i.e.}, without mean-zero and variance decomposition steps), as shown in Figure~\ref{fig:ablation_combined}.

Finally, Figure~\ref{fig:ablation_combined} and Figure~\ref{fig:ablation_heatmap} also demonstrate a degradation in performance as the number of features increases, aligning with the findings in Section~\ref{sec:exp_zero_shot}. Interestingly, TabPFN trained with 10 samples exhibits the opposite trend: its performance typically improves as the number of features grows.

\section{Synthetic-to-Real Distribution Gap}\label{sec:appendix_synthetic_gap}

\begin{figure}[t!]
    \centering
    \includegraphics[width=0.9\linewidth]{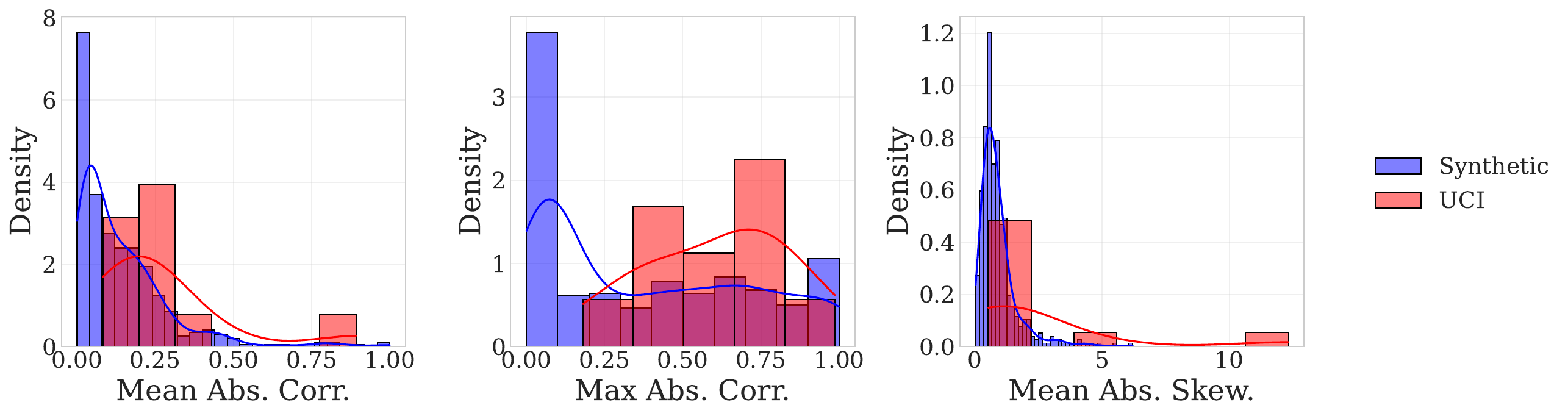}
    \caption{Synthetic-to-real distribution gap. Histogram overlays compare 500 synthetic datasets (blue) against 11 UCI evaluation datasets (red) across three statistics: Mean Abs.\ Corr.\ (mean absolute pairwise correlation), Max Abs.\ Corr.\ (maximum absolute pairwise correlation), and Mean Abs.\ Skew.\ (mean absolute skewness). UCI datasets exhibit higher skewness and sparser correlation structures on average.}
    \label{fig:synthetic_to_real_gap}
\end{figure}

To make the synthetic-to-real gap concrete, we compare statistics of 500 synthetic training datasets (sampled from the same meta-distribution used during pretraining) against the 11 UCI evaluation datasets. Figure~\ref{fig:synthetic_to_real_gap} shows histogram overlays for three statistics: mean absolute pairwise correlation, maximum absolute pairwise correlation, and mean absolute skewness. Table~\ref{tab:synthetic_to_real_gap_summary} summarizes these statistics across synthetic and real datasets.

The synthetic generator produces moderate correlations ($\bar{r} = 0.140$) and moderate skewness ($\bar{s} = 0.911$), whereas UCI datasets show higher correlations ($\bar{r} = 0.272$) and substantially higher skewness ($\bar{s} = 2.347$).

\begin{table}[h]
    \centering
    \caption{Summary statistics comparing synthetic and UCI datasets across three distributional measures.}
    \label{tab:synthetic_to_real_gap_summary}
    \vspace{0.5em}
    \footnotesize
    \begin{tabular}{lrrrrrrrrr}
\toprule
 & \multicolumn{3}{r}{Mean Abs. Corr.} & \multicolumn{3}{r}{Max Abs. Corr.} & \multicolumn{3}{r}{Mean Abs. Skew.} \\
 & mean & std & median & mean & std & median & mean & std & median \\
Dataset &  &  &  &  &  &  &  &  &  \\
\midrule
Synthetic & 0.140 & 0.148 & 0.097 & 0.385 & 0.333 & 0.298 & 0.911 & 0.779 & 0.704 \\
UCI & 0.272 & 0.225 & 0.201 & 0.607 & 0.240 & 0.625 & 2.347 & 3.475 & 0.977 \\
\bottomrule
\end{tabular}

\end{table} These mismatches help explain why ExplainerPFN performs more consistently on synthetic ablations than on real-world benchmarks: the generator does not yet capture the heavy-tailed marginal distributions and sparser correlation structures common in real tabular data. Expanding the generator to sample from richer families of marginal distributions and sparser DAG topologies is a promising direction for improving real-world transfer.

\section{Full Results with Bootstrap Confidence Intervals}\label{sec:appendix_full_tables}

Tables~\ref{tab:results_corr_across_datasets_rfr_with_ci}--\ref{tab:results_spearman_across_datasets_mlp_with_ci} present the full experimental results with bootstrap 95\% confidence intervals for all methods, datasets, and base predictors. Each cell reports the mean correlation and its 95\% CI in parentheses.

\begin{table*}[htb]
    \centering
    \caption{Pearson correlation with 95\% bootstrap confidence intervals for all methods, using a Random Forest base predictor.}
    \label{tab:results_corr_across_datasets_rfr_with_ci}
    \vspace{0.5em}
    \footnotesize
    \resizebox{\textwidth}{!}{\begin{tabular}{ccccccccccccc}
\toprule
Method & Samples & BA & AB & AN & CC & EC & FL & GC & HE & HD & HP & TH \\
\midrule
EPFN & \makecell{0} & \makecell{\underline{0.897} \\ \tiny (0.892--0.904)} & \makecell{0.177 \\ \tiny (0.159--0.189)} & \makecell{0.443 \\ \tiny (0.324--0.528)} & \makecell{\underline{0.479} \\ \tiny (0.456--0.508)} & \makecell{0.397 \\ \tiny (0.298--0.536)} & \makecell{0.098 \\ \tiny (0.036--0.169)} & \makecell{0.160 \\ \tiny (0.128--0.200)} & \makecell{\textbf{0.641} \\ \tiny (0.585--0.709)} & \makecell{\textbf{0.580} \\ \tiny (0.529--0.638)} & \makecell{\textbf{0.643} \\ \tiny (0.510--0.725)} & \makecell{0.317 \\ \tiny (0.296--0.334)} \\
\midrule
 & \makecell{2} & \makecell{-0.058 \\ \tiny (-0.106--0.003)} & \makecell{0.142 \\ \tiny (0.123--0.165)} & \makecell{0.001 \\ \tiny (-0.062--0.064)} & \makecell{-0.013 \\ \tiny (-0.052--0.026)} & \makecell{0.040 \\ \tiny (-0.166--0.239)} & \makecell{0.205 \\ \tiny (0.073--0.317)} & \makecell{0.150 \\ \tiny (0.100--0.189)} & \makecell{-0.073 \\ \tiny (-0.188--0.026)} & \makecell{0.230 \\ \tiny (0.198--0.277)} & \makecell{-0.083 \\ \tiny (-0.204--0.067)} & \makecell{0.053 \\ \tiny (0.008--0.140)} \\
 & \makecell{4} & \makecell{0.019 \\ \tiny (-0.034--0.071)} & \makecell{0.130 \\ \tiny (0.103--0.156)} & \makecell{0.093 \\ \tiny (0.016--0.157)} & \makecell{0.079 \\ \tiny (0.041--0.111)} & \makecell{0.218 \\ \tiny (-0.065--0.403)} & \makecell{0.074 \\ \tiny (-0.065--0.222)} & \makecell{0.159 \\ \tiny (0.112--0.213)} & \makecell{-0.140 \\ \tiny (-0.233---0.045)} & \makecell{0.070 \\ \tiny (0.002--0.135)} & \makecell{-0.081 \\ \tiny (-0.199--0.088)} & \makecell{0.067 \\ \tiny (0.020--0.148)} \\
MLP & \makecell{6} & \makecell{0.065 \\ \tiny (-0.003--0.138)} & \makecell{-0.061 \\ \tiny (-0.076---0.043)} & \makecell{0.091 \\ \tiny (0.014--0.156)} & \makecell{0.075 \\ \tiny (0.030--0.124)} & \makecell{0.320 \\ \tiny (-0.056--0.564)} & \makecell{0.094 \\ \tiny (-0.055--0.231)} & \makecell{0.078 \\ \tiny (0.026--0.135)} & \makecell{-0.132 \\ \tiny (-0.221---0.028)} & \makecell{0.136 \\ \tiny (0.085--0.176)} & \makecell{-0.078 \\ \tiny (-0.206--0.079)} & \makecell{0.065 \\ \tiny (0.018--0.150)} \\
 & \makecell{8} & \makecell{0.656 \\ \tiny (0.629--0.684)} & \makecell{0.023 \\ \tiny (0.001--0.049)} & \makecell{0.088 \\ \tiny (0.014--0.152)} & \makecell{0.090 \\ \tiny (0.047--0.140)} & \makecell{0.316 \\ \tiny (0.007--0.512)} & \makecell{0.129 \\ \tiny (-0.037--0.305)} & \makecell{0.117 \\ \tiny (0.065--0.168)} & \makecell{-0.112 \\ \tiny (-0.202---0.014)} & \makecell{0.204 \\ \tiny (0.169--0.241)} & \makecell{-0.060 \\ \tiny (-0.187--0.096)} & \makecell{0.070 \\ \tiny (0.022--0.155)} \\
 & \makecell{10} & \makecell{0.599 \\ \tiny (0.576--0.627)} & \makecell{-0.065 \\ \tiny (-0.087---0.050)} & \makecell{0.091 \\ \tiny (0.016--0.155)} & \makecell{0.065 \\ \tiny (0.023--0.117)} & \makecell{0.445 \\ \tiny (0.217--0.634)} & \makecell{0.066 \\ \tiny (-0.023--0.146)} & \makecell{0.119 \\ \tiny (0.059--0.164)} & \makecell{-0.091 \\ \tiny (-0.187--0.017)} & \makecell{0.088 \\ \tiny (0.042--0.130)} & \makecell{-0.057 \\ \tiny (-0.184--0.096)} & \makecell{0.062 \\ \tiny (0.013--0.161)} \\
\midrule
 & \makecell{2} & \makecell{0.376 \\ \tiny (0.345--0.405)} & \makecell{0.037 \\ \tiny (0.015--0.058)} & \makecell{0.041 \\ \tiny (-0.030--0.089)} & \makecell{0.134 \\ \tiny (0.095--0.175)} & \makecell{0.304 \\ \tiny (0.005--0.605)} & \makecell{0.121 \\ \tiny (-0.034--0.223)} & \makecell{0.050 \\ \tiny (0.012--0.097)} & \makecell{-0.020 \\ \tiny (-0.135--0.103)} & \makecell{0.220 \\ \tiny (0.162--0.279)} & \makecell{0.413 \\ \tiny (0.325--0.500)} & \makecell{0.035 \\ \tiny (0.010--0.063)} \\
 & \makecell{4} & \makecell{0.556 \\ \tiny (0.530--0.581)} & \makecell{0.093 \\ \tiny (0.059--0.128)} & \makecell{0.176 \\ \tiny (0.108--0.242)} & \makecell{0.251 \\ \tiny (0.200--0.304)} & \makecell{0.520 \\ \tiny (0.300--0.705)} & \makecell{0.427 \\ \tiny (0.290--0.530)} & \makecell{0.182 \\ \tiny (0.151--0.228)} & \makecell{0.260 \\ \tiny (0.145--0.379)} & \makecell{0.287 \\ \tiny (0.230--0.341)} & \makecell{0.434 \\ \tiny (0.326--0.500)} & \makecell{0.095 \\ \tiny (0.064--0.127)} \\
RF & \makecell{6} & \makecell{0.562 \\ \tiny (0.534--0.590)} & \makecell{\textbf{0.377} \\ \tiny (0.355--0.400)} & \makecell{0.272 \\ \tiny (0.205--0.344)} & \makecell{0.332 \\ \tiny (0.292--0.380)} & \makecell{0.767 \\ \tiny (0.661--0.870)} & \makecell{0.561 \\ \tiny (0.455--0.635)} & \makecell{0.283 \\ \tiny (0.227--0.328)} & \makecell{0.359 \\ \tiny (0.282--0.453)} & \makecell{0.315 \\ \tiny (0.267--0.353)} & \makecell{0.340 \\ \tiny (0.216--0.447)} & \makecell{0.310 \\ \tiny (0.280--0.349)} \\
 & \makecell{8} & \makecell{0.767 \\ \tiny (0.751--0.782)} & \makecell{0.274 \\ \tiny (0.245--0.299)} & \makecell{\textbf{0.569} \\ \tiny (0.495--0.661)} & \makecell{0.351 \\ \tiny (0.314--0.395)} & \makecell{0.738 \\ \tiny (0.614--0.863)} & \makecell{\underline{0.577} \\ \tiny (0.475--0.648)} & \makecell{\underline{0.291} \\ \tiny (0.239--0.330)} & \makecell{0.338 \\ \tiny (0.254--0.411)} & \makecell{0.370 \\ \tiny (0.326--0.408)} & \makecell{0.420 \\ \tiny (0.330--0.483)} & \makecell{0.351 \\ \tiny (0.321--0.389)} \\
 & \makecell{10} & \makecell{0.879 \\ \tiny (0.872--0.888)} & \makecell{0.285 \\ \tiny (0.256--0.309)} & \makecell{\underline{0.556} \\ \tiny (0.481--0.649)} & \makecell{0.375 \\ \tiny (0.330--0.418)} & \makecell{\textbf{0.962} \\ \tiny (0.949--0.971)} & \makecell{\textbf{0.596} \\ \tiny (0.496--0.657)} & \makecell{\textbf{0.327} \\ \tiny (0.282--0.364)} & \makecell{0.400 \\ \tiny (0.332--0.457)} & \makecell{0.434 \\ \tiny (0.394--0.466)} & \makecell{0.431 \\ \tiny (0.334--0.486)} & \makecell{\underline{0.594} \\ \tiny (0.567--0.616)} \\
\midrule
 & \makecell{2} & \makecell{0.109 \\ \tiny (0.065--0.140)} & \makecell{0.045 \\ \tiny (0.022--0.068)} & \makecell{-0.120 \\ \tiny (-0.157---0.057)} & \makecell{0.080 \\ \tiny (0.043--0.116)} & \makecell{0.190 \\ \tiny (-0.035--0.453)} & \makecell{0.104 \\ \tiny (-0.016--0.199)} & \makecell{0.065 \\ \tiny (0.025--0.108)} & \makecell{-0.117 \\ \tiny (-0.221--0.007)} & \makecell{-0.056 \\ \tiny (-0.094---0.015)} & \makecell{-0.047 \\ \tiny (-0.174--0.077)} & \makecell{0.034 \\ \tiny (0.010--0.063)} \\
 & \makecell{4} & \makecell{0.668 \\ \tiny (0.655--0.683)} & \makecell{0.040 \\ \tiny (0.015--0.062)} & \makecell{0.226 \\ \tiny (0.160--0.294)} & \makecell{\textbf{0.527} \\ \tiny (0.482--0.565)} & \makecell{0.845 \\ \tiny (0.776--0.896)} & \makecell{0.540 \\ \tiny (0.436--0.621)} & \makecell{0.156 \\ \tiny (0.119--0.197)} & \makecell{0.405 \\ \tiny (0.321--0.493)} & \makecell{0.181 \\ \tiny (0.140--0.222)} & \makecell{0.483 \\ \tiny (0.322--0.625)} & \makecell{0.139 \\ \tiny (0.115--0.166)} \\
TabPFN & \makecell{6} & \makecell{0.738 \\ \tiny (0.722--0.754)} & \makecell{\underline{0.296} \\ \tiny (0.278--0.316)} & \makecell{0.280 \\ \tiny (0.213--0.352)} & \makecell{0.297 \\ \tiny (0.259--0.342)} & \makecell{0.942 \\ \tiny (0.902--0.967)} & \makecell{0.491 \\ \tiny (0.375--0.573)} & \makecell{0.133 \\ \tiny (0.088--0.188)} & \makecell{0.388 \\ \tiny (0.299--0.464)} & \makecell{0.345 \\ \tiny (0.303--0.383)} & \makecell{0.486 \\ \tiny (0.288--0.604)} & \makecell{0.326 \\ \tiny (0.291--0.362)} \\
 & \makecell{8} & \makecell{0.773 \\ \tiny (0.752--0.794)} & \makecell{0.221 \\ \tiny (0.194--0.246)} & \makecell{0.510 \\ \tiny (0.434--0.606)} & \makecell{0.345 \\ \tiny (0.311--0.386)} & \makecell{0.952 \\ \tiny (0.906--0.973)} & \makecell{0.341 \\ \tiny (0.226--0.428)} & \makecell{0.245 \\ \tiny (0.187--0.282)} & \makecell{0.445 \\ \tiny (0.365--0.502)} & \makecell{0.388 \\ \tiny (0.347--0.425)} & \makecell{\underline{0.546} \\ \tiny (0.375--0.653)} & \makecell{0.404 \\ \tiny (0.372--0.443)} \\
 & \makecell{10} & \makecell{\textbf{0.903} \\ \tiny (0.896--0.911)} & \makecell{0.232 \\ \tiny (0.204--0.256)} & \makecell{0.517 \\ \tiny (0.447--0.619)} & \makecell{0.446 \\ \tiny (0.413--0.487)} & \makecell{\underline{0.962} \\ \tiny (0.921--0.983)} & \makecell{0.531 \\ \tiny (0.415--0.632)} & \makecell{0.247 \\ \tiny (0.201--0.282)} & \makecell{\underline{0.448} \\ \tiny (0.367--0.507)} & \makecell{\underline{0.459} \\ \tiny (0.412--0.498)} & \makecell{0.531 \\ \tiny (0.389--0.632)} & \makecell{\textbf{0.732} \\ \tiny (0.713--0.752)} \\
\bottomrule
\end{tabular}}
\end{table*}

\begin{table*}[htb]
    \centering
    \caption{Spearman rank correlation with 95\% bootstrap confidence intervals for all methods, using a Random Forest base predictor.}
    \label{tab:results_spearman_across_datasets_rfr_with_ci}
    \vspace{0.5em}
    \footnotesize
    \resizebox{\textwidth}{!}{\begin{tabular}{ccccccccccccc}
\toprule
Method & Samples & BA & AB & AN & CC & EC & FL & GC & HE & HD & HP & TH \\
\midrule
EPFN & \makecell{0} & \makecell{\textbf{0.923} \\ \tiny (0.918--0.929)} & \makecell{\textbf{0.373} \\ \tiny (0.351--0.388)} & \makecell{0.298 \\ \tiny (0.255--0.349)} & \makecell{\textbf{0.518} \\ \tiny (0.497--0.547)} & \makecell{0.373 \\ \tiny (0.255--0.483)} & \makecell{0.235 \\ \tiny (0.167--0.330)} & \makecell{0.115 \\ \tiny (0.078--0.166)} & \makecell{\textbf{0.573} \\ \tiny (0.522--0.630)} & \makecell{\textbf{0.466} \\ \tiny (0.423--0.515)} & \makecell{0.434 \\ \tiny (0.345--0.519)} & \makecell{\textbf{0.672} \\ \tiny (0.660--0.688)} \\
\midrule
 & \makecell{2} & \makecell{-0.002 \\ \tiny (-0.054--0.036)} & \makecell{0.144 \\ \tiny (0.127--0.163)} & \makecell{-0.047 \\ \tiny (-0.124--0.015)} & \makecell{-0.020 \\ \tiny (-0.070--0.018)} & \makecell{0.112 \\ \tiny (-0.060--0.293)} & \makecell{0.251 \\ \tiny (0.111--0.363)} & \makecell{0.118 \\ \tiny (0.057--0.158)} & \makecell{-0.065 \\ \tiny (-0.174--0.028)} & \makecell{0.240 \\ \tiny (0.201--0.296)} & \makecell{-0.112 \\ \tiny (-0.226---0.016)} & \makecell{0.058 \\ \tiny (0.031--0.083)} \\
 & \makecell{4} & \makecell{0.105 \\ \tiny (0.053--0.157)} & \makecell{0.176 \\ \tiny (0.147--0.199)} & \makecell{0.051 \\ \tiny (-0.014--0.120)} & \makecell{0.099 \\ \tiny (0.055--0.134)} & \makecell{0.163 \\ \tiny (-0.132--0.399)} & \makecell{0.127 \\ \tiny (-0.028--0.278)} & \makecell{0.085 \\ \tiny (0.049--0.131)} & \makecell{-0.140 \\ \tiny (-0.224---0.038)} & \makecell{0.256 \\ \tiny (0.190--0.306)} & \makecell{-0.107 \\ \tiny (-0.249--0.016)} & \makecell{0.087 \\ \tiny (0.057--0.112)} \\
MLP & \makecell{6} & \makecell{0.236 \\ \tiny (0.189--0.298)} & \makecell{-0.105 \\ \tiny (-0.125---0.082)} & \makecell{-0.023 \\ \tiny (-0.090--0.037)} & \makecell{0.092 \\ \tiny (0.047--0.157)} & \makecell{0.184 \\ \tiny (-0.135--0.412)} & \makecell{0.077 \\ \tiny (-0.085--0.271)} & \makecell{0.033 \\ \tiny (-0.009--0.087)} & \makecell{-0.143 \\ \tiny (-0.220---0.052)} & \makecell{0.229 \\ \tiny (0.174--0.290)} & \makecell{-0.070 \\ \tiny (-0.194--0.035)} & \makecell{0.089 \\ \tiny (0.059--0.116)} \\
 & \makecell{8} & \makecell{0.687 \\ \tiny (0.666--0.718)} & \makecell{-0.019 \\ \tiny (-0.041--0.004)} & \makecell{0.107 \\ \tiny (0.024--0.172)} & \makecell{0.109 \\ \tiny (0.064--0.155)} & \makecell{0.173 \\ \tiny (-0.121--0.367)} & \makecell{0.122 \\ \tiny (-0.042--0.315)} & \makecell{0.064 \\ \tiny (0.005--0.119)} & \makecell{-0.093 \\ \tiny (-0.190--0.005)} & \makecell{0.200 \\ \tiny (0.152--0.246)} & \makecell{-0.039 \\ \tiny (-0.176--0.078)} & \makecell{0.113 \\ \tiny (0.085--0.138)} \\
 & \makecell{10} & \makecell{0.632 \\ \tiny (0.608--0.660)} & \makecell{-0.117 \\ \tiny (-0.141---0.095)} & \makecell{0.119 \\ \tiny (0.043--0.177)} & \makecell{0.078 \\ \tiny (0.037--0.131)} & \makecell{0.212 \\ \tiny (0.038--0.407)} & \makecell{0.052 \\ \tiny (-0.052--0.174)} & \makecell{0.076 \\ \tiny (0.023--0.132)} & \makecell{-0.076 \\ \tiny (-0.177--0.032)} & \makecell{0.140 \\ \tiny (0.088--0.197)} & \makecell{-0.024 \\ \tiny (-0.159--0.101)} & \makecell{0.066 \\ \tiny (0.040--0.092)} \\
\midrule
 & \makecell{2} & \makecell{0.326 \\ \tiny (0.287--0.366)} & \makecell{0.036 \\ \tiny (0.012--0.059)} & \makecell{0.206 \\ \tiny (0.138--0.253)} & \makecell{0.214 \\ \tiny (0.178--0.247)} & \makecell{-0.018 \\ \tiny (-0.208--0.223)} & \makecell{-0.014 \\ \tiny (-0.065--0.031)} & \makecell{0.105 \\ \tiny (0.067--0.149)} & \makecell{-0.072 \\ \tiny (-0.161--0.011)} & \makecell{0.206 \\ \tiny (0.137--0.287)} & \makecell{\textbf{0.472} \\ \tiny (0.366--0.579)} & \makecell{-0.018 \\ \tiny (-0.032--0.004)} \\
 & \makecell{4} & \makecell{0.468 \\ \tiny (0.431--0.503)} & \makecell{0.108 \\ \tiny (0.080--0.137)} & \makecell{0.429 \\ \tiny (0.376--0.472)} & \makecell{0.337 \\ \tiny (0.299--0.376)} & \makecell{0.325 \\ \tiny (0.084--0.505)} & \makecell{0.267 \\ \tiny (0.203--0.339)} & \makecell{0.286 \\ \tiny (0.251--0.329)} & \makecell{0.163 \\ \tiny (0.078--0.264)} & \makecell{0.270 \\ \tiny (0.196--0.330)} & \makecell{\underline{0.462} \\ \tiny (0.344--0.565)} & \makecell{0.017 \\ \tiny (0.005--0.037)} \\
RF & \makecell{6} & \makecell{0.522 \\ \tiny (0.490--0.553)} & \makecell{\underline{0.361} \\ \tiny (0.344--0.377)} & \makecell{0.535 \\ \tiny (0.489--0.588)} & \makecell{0.356 \\ \tiny (0.322--0.393)} & \makecell{0.517 \\ \tiny (0.342--0.618)} & \makecell{0.451 \\ \tiny (0.361--0.533)} & \makecell{\textbf{0.392} \\ \tiny (0.351--0.429)} & \makecell{0.273 \\ \tiny (0.198--0.350)} & \makecell{0.356 \\ \tiny (0.297--0.403)} & \makecell{0.298 \\ \tiny (0.156--0.435)} & \makecell{0.175 \\ \tiny (0.162--0.195)} \\
 & \makecell{8} & \makecell{0.768 \\ \tiny (0.744--0.794)} & \makecell{0.322 \\ \tiny (0.298--0.349)} & \makecell{\textbf{0.578} \\ \tiny (0.529--0.626)} & \makecell{0.365 \\ \tiny (0.332--0.405)} & \makecell{0.614 \\ \tiny (0.455--0.698)} & \makecell{0.519 \\ \tiny (0.436--0.601)} & \makecell{0.341 \\ \tiny (0.313--0.378)} & \makecell{0.223 \\ \tiny (0.132--0.277)} & \makecell{0.389 \\ \tiny (0.340--0.428)} & \makecell{0.445 \\ \tiny (0.317--0.538)} & \makecell{0.332 \\ \tiny (0.310--0.355)} \\
 & \makecell{10} & \makecell{\underline{0.911} \\ \tiny (0.901--0.922)} & \makecell{0.330 \\ \tiny (0.305--0.356)} & \makecell{\underline{0.557} \\ \tiny (0.505--0.600)} & \makecell{0.411 \\ \tiny (0.380--0.439)} & \makecell{\textbf{0.805} \\ \tiny (0.751--0.852)} & \makecell{\textbf{0.594} \\ \tiny (0.510--0.663)} & \makecell{\underline{0.377} \\ \tiny (0.349--0.406)} & \makecell{0.341 \\ \tiny (0.269--0.401)} & \makecell{\underline{0.449} \\ \tiny (0.398--0.489)} & \makecell{0.454 \\ \tiny (0.329--0.549)} & \makecell{0.427 \\ \tiny (0.410--0.443)} \\
\midrule
 & \makecell{2} & \makecell{0.054 \\ \tiny (0.007--0.098)} & \makecell{0.059 \\ \tiny (0.037--0.078)} & \makecell{0.035 \\ \tiny (-0.015--0.097)} & \makecell{0.199 \\ \tiny (0.160--0.224)} & \makecell{-0.045 \\ \tiny (-0.246--0.198)} & \makecell{0.009 \\ \tiny (-0.051--0.062)} & \makecell{0.153 \\ \tiny (0.115--0.194)} & \makecell{-0.127 \\ \tiny (-0.232---0.055)} & \makecell{-0.062 \\ \tiny (-0.115---0.014)} & \makecell{0.081 \\ \tiny (-0.054--0.232)} & \makecell{-0.009 \\ \tiny (-0.022--0.008)} \\
 & \makecell{4} & \makecell{0.783 \\ \tiny (0.768--0.800)} & \makecell{0.082 \\ \tiny (0.057--0.106)} & \makecell{0.460 \\ \tiny (0.409--0.503)} & \makecell{\underline{0.518} \\ \tiny (0.492--0.552)} & \makecell{0.460 \\ \tiny (0.248--0.623)} & \makecell{0.392 \\ \tiny (0.319--0.461)} & \makecell{0.220 \\ \tiny (0.183--0.260)} & \makecell{0.283 \\ \tiny (0.224--0.361)} & \makecell{0.277 \\ \tiny (0.214--0.324)} & \makecell{0.205 \\ \tiny (0.011--0.369)} & \makecell{0.144 \\ \tiny (0.128--0.162)} \\
TabPFN & \makecell{6} & \makecell{0.697 \\ \tiny (0.678--0.718)} & \makecell{0.320 \\ \tiny (0.306--0.340)} & \makecell{0.506 \\ \tiny (0.457--0.551)} & \makecell{0.329 \\ \tiny (0.293--0.367)} & \makecell{0.621 \\ \tiny (0.527--0.688)} & \makecell{0.460 \\ \tiny (0.363--0.532)} & \makecell{0.250 \\ \tiny (0.204--0.292)} & \makecell{0.294 \\ \tiny (0.217--0.371)} & \makecell{0.360 \\ \tiny (0.306--0.410)} & \makecell{0.193 \\ \tiny (0.024--0.368)} & \makecell{0.437 \\ \tiny (0.420--0.452)} \\
 & \makecell{8} & \makecell{0.754 \\ \tiny (0.722--0.786)} & \makecell{0.312 \\ \tiny (0.287--0.344)} & \makecell{0.542 \\ \tiny (0.486--0.603)} & \makecell{0.341 \\ \tiny (0.309--0.379)} & \makecell{0.747 \\ \tiny (0.651--0.810)} & \makecell{0.447 \\ \tiny (0.375--0.536)} & \makecell{0.357 \\ \tiny (0.324--0.400)} & \makecell{0.326 \\ \tiny (0.262--0.368)} & \makecell{0.412 \\ \tiny (0.368--0.448)} & \makecell{0.198 \\ \tiny (0.012--0.351)} & \makecell{0.627 \\ \tiny (0.617--0.639)} \\
 & \makecell{10} & \makecell{0.896 \\ \tiny (0.882--0.906)} & \makecell{0.314 \\ \tiny (0.287--0.345)} & \makecell{0.557 \\ \tiny (0.506--0.617)} & \makecell{0.459 \\ \tiny (0.435--0.495)} & \makecell{\underline{0.788} \\ \tiny (0.703--0.844)} & \makecell{\underline{0.555} \\ \tiny (0.467--0.645)} & \makecell{0.302 \\ \tiny (0.274--0.345)} & \makecell{\underline{0.392} \\ \tiny (0.329--0.447)} & \makecell{0.427 \\ \tiny (0.383--0.475)} & \makecell{0.268 \\ \tiny (0.114--0.384)} & \makecell{\underline{0.637} \\ \tiny (0.625--0.651)} \\
\bottomrule
\end{tabular}}
\end{table*}

\begin{table*}[htb]
    \centering
    \caption{Pearson correlation with 95\% bootstrap confidence intervals for all methods, using an MLP base predictor.}
    \label{tab:results_corr_across_datasets_mlp_with_ci}
    \vspace{0.5em}
    \footnotesize
    \resizebox{\textwidth}{!}{\begin{tabular}{ccccccccccccc}
\toprule
Method & Samples & BA & AB & AN & CC & EC & FL & GC & HE & HD & HP & TH \\
\midrule
EPFN & \makecell{0} & \makecell{0.880 \\ \tiny (0.872--0.886)} & \makecell{0.114 \\ \tiny (0.084--0.146)} & \makecell{\textbf{0.338} \\ \tiny (0.183--0.466)} & \makecell{0.530 \\ \tiny (0.510--0.552)} & \makecell{0.490 \\ \tiny (0.263--0.605)} & \makecell{0.058 \\ \tiny (-0.011--0.123)} & \makecell{0.213 \\ \tiny (0.161--0.254)} & \makecell{0.463 \\ \tiny (0.402--0.511)} & \makecell{\textbf{0.557} \\ \tiny (0.515--0.604)} & \makecell{\textbf{0.485} \\ \tiny (0.223--0.691)} & \makecell{0.212 \\ \tiny (0.183--0.232)} \\
\midrule
 & \makecell{2} & \makecell{-0.052 \\ \tiny (-0.099--0.010)} & \makecell{-0.005 \\ \tiny (-0.019--0.007)} & \makecell{-0.046 \\ \tiny (-0.097--0.002)} & \makecell{0.032 \\ \tiny (-0.011--0.073)} & \makecell{0.064 \\ \tiny (-0.125--0.210)} & \makecell{0.102 \\ \tiny (-0.027--0.229)} & \makecell{0.178 \\ \tiny (0.141--0.221)} & \makecell{-0.131 \\ \tiny (-0.244---0.039)} & \makecell{0.210 \\ \tiny (0.167--0.247)} & \makecell{-0.023 \\ \tiny (-0.228--0.196)} & \makecell{0.045 \\ \tiny (0.009--0.115)} \\
 & \makecell{4} & \makecell{0.027 \\ \tiny (-0.028--0.081)} & \makecell{0.221 \\ \tiny (0.202--0.242)} & \makecell{0.153 \\ \tiny (0.037--0.251)} & \makecell{0.067 \\ \tiny (0.026--0.102)} & \makecell{0.216 \\ \tiny (-0.046--0.395)} & \makecell{0.068 \\ \tiny (-0.089--0.279)} & \makecell{0.157 \\ \tiny (0.098--0.217)} & \makecell{-0.170 \\ \tiny (-0.260---0.083)} & \makecell{0.053 \\ \tiny (-0.011--0.120)} & \makecell{0.056 \\ \tiny (-0.127--0.250)} & \makecell{0.050 \\ \tiny (0.017--0.110)} \\
MLP & \makecell{6} & \makecell{0.088 \\ \tiny (0.019--0.166)} & \makecell{0.155 \\ \tiny (0.137--0.177)} & \makecell{0.151 \\ \tiny (0.035--0.250)} & \makecell{0.051 \\ \tiny (0.012--0.092)} & \makecell{0.363 \\ \tiny (0.033--0.602)} & \makecell{0.147 \\ \tiny (0.035--0.256)} & \makecell{0.104 \\ \tiny (0.045--0.186)} & \makecell{-0.151 \\ \tiny (-0.244---0.058)} & \makecell{0.119 \\ \tiny (0.068--0.177)} & \makecell{0.039 \\ \tiny (-0.135--0.244)} & \makecell{0.053 \\ \tiny (0.016--0.123)} \\
 & \makecell{8} & \makecell{0.744 \\ \tiny (0.721--0.770)} & \makecell{0.180 \\ \tiny (0.151--0.204)} & \makecell{0.150 \\ \tiny (0.034--0.250)} & \makecell{0.076 \\ \tiny (0.041--0.112)} & \makecell{0.329 \\ \tiny (0.052--0.553)} & \makecell{0.173 \\ \tiny (0.056--0.280)} & \makecell{0.163 \\ \tiny (0.097--0.234)} & \makecell{-0.136 \\ \tiny (-0.232---0.053)} & \makecell{0.192 \\ \tiny (0.145--0.238)} & \makecell{0.054 \\ \tiny (-0.120--0.254)} & \makecell{0.057 \\ \tiny (0.020--0.128)} \\
 & \makecell{10} & \makecell{0.734 \\ \tiny (0.702--0.753)} & \makecell{0.144 \\ \tiny (0.127--0.158)} & \makecell{0.151 \\ \tiny (0.034--0.250)} & \makecell{0.074 \\ \tiny (0.038--0.112)} & \makecell{0.466 \\ \tiny (0.273--0.639)} & \makecell{0.167 \\ \tiny (0.094--0.240)} & \makecell{0.143 \\ \tiny (0.092--0.195)} & \makecell{-0.127 \\ \tiny (-0.212---0.033)} & \makecell{0.080 \\ \tiny (0.027--0.132)} & \makecell{0.055 \\ \tiny (-0.116--0.253)} & \makecell{0.055 \\ \tiny (0.012--0.141)} \\
\midrule
 & \makecell{2} & \makecell{0.207 \\ \tiny (0.164--0.252)} & \makecell{0.021 \\ \tiny (-0.016--0.050)} & \makecell{0.030 \\ \tiny (-0.025--0.077)} & \makecell{0.166 \\ \tiny (0.098--0.228)} & \makecell{0.195 \\ \tiny (-0.018--0.365)} & \makecell{0.098 \\ \tiny (-0.046--0.217)} & \makecell{0.055 \\ \tiny (0.011--0.095)} & \makecell{-0.119 \\ \tiny (-0.213---0.025)} & \makecell{0.172 \\ \tiny (0.110--0.224)} & \makecell{-0.002 \\ \tiny (-0.147--0.151)} & \makecell{0.053 \\ \tiny (0.030--0.082)} \\
 & \makecell{4} & \makecell{0.364 \\ \tiny (0.319--0.406)} & \makecell{0.092 \\ \tiny (0.048--0.123)} & \makecell{0.051 \\ \tiny (-0.007--0.097)} & \makecell{0.312 \\ \tiny (0.257--0.371)} & \makecell{0.362 \\ \tiny (0.095--0.563)} & \makecell{0.250 \\ \tiny (0.094--0.410)} & \makecell{0.221 \\ \tiny (0.184--0.263)} & \makecell{0.158 \\ \tiny (0.062--0.285)} & \makecell{0.213 \\ \tiny (0.152--0.271)} & \makecell{0.039 \\ \tiny (-0.139--0.185)} & \makecell{0.121 \\ \tiny (0.092--0.152)} \\
RF & \makecell{6} & \makecell{0.437 \\ \tiny (0.398--0.476)} & \makecell{0.628 \\ \tiny (0.611--0.646)} & \makecell{0.172 \\ \tiny (0.124--0.213)} & \makecell{0.273 \\ \tiny (0.215--0.331)} & \makecell{0.734 \\ \tiny (0.606--0.829)} & \makecell{0.374 \\ \tiny (0.247--0.493)} & \makecell{0.311 \\ \tiny (0.261--0.366)} & \makecell{0.242 \\ \tiny (0.158--0.336)} & \makecell{0.259 \\ \tiny (0.195--0.309)} & \makecell{0.025 \\ \tiny (-0.112--0.146)} & \makecell{0.342 \\ \tiny (0.314--0.383)} \\
 & \makecell{8} & \makecell{0.775 \\ \tiny (0.758--0.788)} & \makecell{\textbf{0.687} \\ \tiny (0.670--0.700)} & \makecell{0.263 \\ \tiny (0.214--0.316)} & \makecell{0.453 \\ \tiny (0.406--0.500)} & \makecell{0.682 \\ \tiny (0.515--0.805)} & \makecell{0.416 \\ \tiny (0.308--0.520)} & \makecell{0.342 \\ \tiny (0.302--0.387)} & \makecell{0.320 \\ \tiny (0.244--0.408)} & \makecell{0.386 \\ \tiny (0.327--0.430)} & \makecell{0.046 \\ \tiny (-0.095--0.175)} & \makecell{0.373 \\ \tiny (0.344--0.413)} \\
 & \makecell{10} & \makecell{0.830 \\ \tiny (0.821--0.839)} & \makecell{\underline{0.684} \\ \tiny (0.667--0.697)} & \makecell{\underline{0.290} \\ \tiny (0.242--0.344)} & \makecell{0.561 \\ \tiny (0.512--0.600)} & \makecell{0.909 \\ \tiny (0.874--0.941)} & \makecell{\underline{0.462} \\ \tiny (0.359--0.577)} & \makecell{0.355 \\ \tiny (0.329--0.387)} & \makecell{0.386 \\ \tiny (0.327--0.451)} & \makecell{\underline{0.456} \\ \tiny (0.420--0.499)} & \makecell{0.044 \\ \tiny (-0.099--0.165)} & \makecell{\underline{0.608} \\ \tiny (0.585--0.632)} \\
\midrule
 & \makecell{2} & \makecell{0.073 \\ \tiny (0.026--0.120)} & \makecell{0.022 \\ \tiny (-0.014--0.051)} & \makecell{0.033 \\ \tiny (-0.023--0.081)} & \makecell{0.061 \\ \tiny (0.010--0.114)} & \makecell{0.074 \\ \tiny (-0.090--0.208)} & \makecell{0.054 \\ \tiny (-0.051--0.154)} & \makecell{0.049 \\ \tiny (0.007--0.088)} & \makecell{-0.145 \\ \tiny (-0.233---0.046)} & \makecell{0.008 \\ \tiny (-0.047--0.066)} & \makecell{-0.013 \\ \tiny (-0.157--0.128)} & \makecell{0.050 \\ \tiny (0.029--0.076)} \\
 & \makecell{4} & \makecell{0.714 \\ \tiny (0.693--0.736)} & \makecell{0.149 \\ \tiny (0.119--0.169)} & \makecell{0.020 \\ \tiny (-0.045--0.071)} & \makecell{\underline{0.632} \\ \tiny (0.595--0.660)} & \makecell{0.808 \\ \tiny (0.714--0.897)} & \makecell{0.396 \\ \tiny (0.302--0.514)} & \makecell{0.282 \\ \tiny (0.243--0.332)} & \makecell{0.163 \\ \tiny (0.062--0.310)} & \makecell{0.328 \\ \tiny (0.291--0.368)} & \makecell{0.054 \\ \tiny (-0.064--0.210)} & \makecell{0.143 \\ \tiny (0.117--0.171)} \\
TabPFN & \makecell{6} & \makecell{0.837 \\ \tiny (0.828--0.853)} & \makecell{0.599 \\ \tiny (0.580--0.615)} & \makecell{0.064 \\ \tiny (0.007--0.119)} & \makecell{0.439 \\ \tiny (0.385--0.486)} & \makecell{0.902 \\ \tiny (0.862--0.929)} & \makecell{0.461 \\ \tiny (0.342--0.582)} & \makecell{0.258 \\ \tiny (0.215--0.316)} & \makecell{0.392 \\ \tiny (0.329--0.477)} & \makecell{0.358 \\ \tiny (0.306--0.403)} & \makecell{0.078 \\ \tiny (-0.051--0.218)} & \makecell{0.329 \\ \tiny (0.296--0.367)} \\
 & \makecell{8} & \makecell{\underline{0.903} \\ \tiny (0.893--0.912)} & \makecell{0.607 \\ \tiny (0.584--0.624)} & \makecell{0.156 \\ \tiny (0.097--0.215)} & \makecell{0.515 \\ \tiny (0.464--0.561)} & \makecell{\underline{0.912} \\ \tiny (0.868--0.939)} & \makecell{0.450 \\ \tiny (0.345--0.554)} & \makecell{\underline{0.366} \\ \tiny (0.328--0.413)} & \makecell{\underline{0.513} \\ \tiny (0.459--0.561)} & \makecell{0.428 \\ \tiny (0.378--0.467)} & \makecell{\underline{0.101} \\ \tiny (-0.001--0.241)} & \makecell{0.346 \\ \tiny (0.312--0.383)} \\
 & \makecell{10} & \makecell{\textbf{0.956} \\ \tiny (0.952--0.959)} & \makecell{0.643 \\ \tiny (0.626--0.658)} & \makecell{0.167 \\ \tiny (0.108--0.221)} & \makecell{\textbf{0.790} \\ \tiny (0.766--0.808)} & \makecell{\textbf{0.925} \\ \tiny (0.878--0.949)} & \makecell{\textbf{0.500} \\ \tiny (0.394--0.617)} & \makecell{\textbf{0.445} \\ \tiny (0.405--0.473)} & \makecell{\textbf{0.538} \\ \tiny (0.475--0.592)} & \makecell{0.214 \\ \tiny (0.155--0.256)} & \makecell{0.088 \\ \tiny (-0.054--0.223)} & \makecell{\textbf{0.693} \\ \tiny (0.673--0.712)} \\
\bottomrule
\end{tabular}}
\end{table*}

\begin{table*}[htb]
    \centering
    \caption{Spearman rank correlation with 95\% bootstrap confidence intervals for all methods, using an MLP base predictor.}
    \label{tab:results_spearman_across_datasets_mlp_with_ci}
    \vspace{0.5em}
    \footnotesize
    \resizebox{\textwidth}{!}{\begin{tabular}{ccccccccccccc}
\toprule
Method & Samples & BA & AB & AN & CC & EC & FL & GC & HE & HD & HP & TH \\
\midrule
EPFN & \makecell{0} & \makecell{0.920 \\ \tiny (0.914--0.929)} & \makecell{0.185 \\ \tiny (0.162--0.208)} & \makecell{0.306 \\ \tiny (0.244--0.345)} & \makecell{0.522 \\ \tiny (0.500--0.556)} & \makecell{0.534 \\ \tiny (0.367--0.644)} & \makecell{0.244 \\ \tiny (0.169--0.327)} & \makecell{0.211 \\ \tiny (0.168--0.252)} & \makecell{\underline{0.405} \\ \tiny (0.343--0.474)} & \makecell{\underline{0.466} \\ \tiny (0.418--0.501)} & \makecell{-0.182 \\ \tiny (-0.350---0.021)} & \makecell{0.134 \\ \tiny (0.121--0.145)} \\
\midrule
 & \makecell{2} & \makecell{0.059 \\ \tiny (0.007--0.098)} & \makecell{0.089 \\ \tiny (0.075--0.102)} & \makecell{0.002 \\ \tiny (-0.053--0.054)} & \makecell{0.030 \\ \tiny (-0.007--0.065)} & \makecell{-0.019 \\ \tiny (-0.202--0.158)} & \makecell{0.212 \\ \tiny (0.120--0.285)} & \makecell{0.179 \\ \tiny (0.133--0.222)} & \makecell{-0.162 \\ \tiny (-0.266---0.049)} & \makecell{0.199 \\ \tiny (0.160--0.238)} & \makecell{-0.017 \\ \tiny (-0.231--0.179)} & \makecell{0.005 \\ \tiny (-0.014--0.030)} \\
 & \makecell{4} & \makecell{0.178 \\ \tiny (0.119--0.239)} & \makecell{0.231 \\ \tiny (0.219--0.246)} & \makecell{0.021 \\ \tiny (-0.054--0.086)} & \makecell{0.093 \\ \tiny (0.060--0.127)} & \makecell{0.051 \\ \tiny (-0.153--0.217)} & \makecell{0.111 \\ \tiny (-0.076--0.240)} & \makecell{0.093 \\ \tiny (0.042--0.130)} & \makecell{-0.189 \\ \tiny (-0.269---0.098)} & \makecell{0.245 \\ \tiny (0.181--0.304)} & \makecell{0.196 \\ \tiny (0.038--0.338)} & \makecell{0.047 \\ \tiny (0.023--0.065)} \\
MLP & \makecell{6} & \makecell{0.324 \\ \tiny (0.275--0.391)} & \makecell{0.053 \\ \tiny (0.035--0.074)} & \makecell{-0.002 \\ \tiny (-0.075--0.057)} & \makecell{0.095 \\ \tiny (0.071--0.123)} & \makecell{0.119 \\ \tiny (-0.121--0.327)} & \makecell{0.085 \\ \tiny (-0.091--0.221)} & \makecell{0.030 \\ \tiny (-0.007--0.075)} & \makecell{-0.176 \\ \tiny (-0.262---0.064)} & \makecell{0.214 \\ \tiny (0.161--0.269)} & \makecell{0.160 \\ \tiny (0.008--0.323)} & \makecell{0.057 \\ \tiny (0.033--0.075)} \\
 & \makecell{8} & \makecell{0.785 \\ \tiny (0.766--0.803)} & \makecell{0.192 \\ \tiny (0.167--0.217)} & \makecell{0.041 \\ \tiny (-0.023--0.106)} & \makecell{0.126 \\ \tiny (0.092--0.166)} & \makecell{0.113 \\ \tiny (-0.057--0.253)} & \makecell{0.143 \\ \tiny (-0.009--0.290)} & \makecell{0.087 \\ \tiny (0.039--0.128)} & \makecell{-0.146 \\ \tiny (-0.231---0.059)} & \makecell{0.188 \\ \tiny (0.130--0.238)} & \makecell{0.179 \\ \tiny (0.009--0.344)} & \makecell{0.070 \\ \tiny (0.044--0.089)} \\
 & \makecell{10} & \makecell{0.739 \\ \tiny (0.704--0.760)} & \makecell{0.122 \\ \tiny (0.106--0.139)} & \makecell{0.057 \\ \tiny (-0.006--0.122)} & \makecell{0.100 \\ \tiny (0.073--0.140)} & \makecell{0.228 \\ \tiny (0.058--0.417)} & \makecell{0.170 \\ \tiny (0.052--0.267)} & \makecell{0.084 \\ \tiny (0.036--0.132)} & \makecell{-0.159 \\ \tiny (-0.249---0.080)} & \makecell{0.120 \\ \tiny (0.072--0.172)} & \makecell{0.176 \\ \tiny (-0.000--0.343)} & \makecell{0.006 \\ \tiny (-0.012--0.025)} \\
\midrule
 & \makecell{2} & \makecell{0.298 \\ \tiny (0.242--0.341)} & \makecell{0.037 \\ \tiny (-0.002--0.068)} & \makecell{0.300 \\ \tiny (0.229--0.357)} & \makecell{0.288 \\ \tiny (0.248--0.321)} & \makecell{0.122 \\ \tiny (-0.061--0.321)} & \makecell{0.047 \\ \tiny (-0.043--0.152)} & \makecell{0.122 \\ \tiny (0.086--0.161)} & \makecell{-0.189 \\ \tiny (-0.252---0.108)} & \makecell{0.194 \\ \tiny (0.109--0.256)} & \makecell{0.067 \\ \tiny (-0.083--0.222)} & \makecell{0.042 \\ \tiny (0.023--0.060)} \\
 & \makecell{4} & \makecell{0.387 \\ \tiny (0.339--0.428)} & \makecell{0.097 \\ \tiny (0.052--0.134)} & \makecell{0.359 \\ \tiny (0.298--0.419)} & \makecell{0.227 \\ \tiny (0.176--0.282)} & \makecell{0.223 \\ \tiny (0.060--0.405)} & \makecell{0.224 \\ \tiny (0.138--0.294)} & \makecell{0.186 \\ \tiny (0.145--0.226)} & \makecell{0.023 \\ \tiny (-0.077--0.137)} & \makecell{0.219 \\ \tiny (0.150--0.277)} & \makecell{0.293 \\ \tiny (0.169--0.398)} & \makecell{0.132 \\ \tiny (0.114--0.146)} \\
RF & \makecell{6} & \makecell{0.499 \\ \tiny (0.457--0.531)} & \makecell{0.614 \\ \tiny (0.595--0.634)} & \makecell{0.481 \\ \tiny (0.430--0.528)} & \makecell{0.275 \\ \tiny (0.216--0.330)} & \makecell{0.479 \\ \tiny (0.344--0.577)} & \makecell{0.352 \\ \tiny (0.243--0.432)} & \makecell{\underline{0.351} \\ \tiny (0.305--0.397)} & \makecell{0.121 \\ \tiny (0.042--0.228)} & \makecell{0.275 \\ \tiny (0.208--0.320)} & \makecell{0.211 \\ \tiny (0.056--0.309)} & \makecell{0.293 \\ \tiny (0.276--0.310)} \\
 & \makecell{8} & \makecell{0.808 \\ \tiny (0.791--0.823)} & \makecell{0.689 \\ \tiny (0.673--0.705)} & \makecell{\underline{0.550} \\ \tiny (0.517--0.593)} & \makecell{0.521 \\ \tiny (0.480--0.559)} & \makecell{0.508 \\ \tiny (0.379--0.618)} & \makecell{0.515 \\ \tiny (0.429--0.584)} & \makecell{0.303 \\ \tiny (0.260--0.345)} & \makecell{0.251 \\ \tiny (0.177--0.323)} & \makecell{0.392 \\ \tiny (0.330--0.440)} & \makecell{0.257 \\ \tiny (0.122--0.352)} & \makecell{0.396 \\ \tiny (0.376--0.412)} \\
 & \makecell{10} & \makecell{0.868 \\ \tiny (0.857--0.878)} & \makecell{\underline{0.690} \\ \tiny (0.674--0.706)} & \makecell{\textbf{0.571} \\ \tiny (0.541--0.614)} & \makecell{\underline{0.599} \\ \tiny (0.569--0.631)} & \makecell{\textbf{0.696} \\ \tiny (0.615--0.770)} & \makecell{\underline{0.607} \\ \tiny (0.534--0.664)} & \makecell{\textbf{0.353} \\ \tiny (0.318--0.392)} & \makecell{0.319 \\ \tiny (0.260--0.380)} & \makecell{\textbf{0.475} \\ \tiny (0.437--0.518)} & \makecell{0.286 \\ \tiny (0.169--0.384)} & \makecell{0.577 \\ \tiny (0.563--0.591)} \\
\midrule
 & \makecell{2} & \makecell{0.104 \\ \tiny (0.053--0.151)} & \makecell{0.030 \\ \tiny (-0.009--0.060)} & \makecell{0.319 \\ \tiny (0.246--0.382)} & \makecell{0.168 \\ \tiny (0.135--0.200)} & \makecell{0.097 \\ \tiny (-0.097--0.283)} & \makecell{0.014 \\ \tiny (-0.062--0.098)} & \makecell{0.110 \\ \tiny (0.071--0.146)} & \makecell{-0.182 \\ \tiny (-0.239---0.108)} & \makecell{-0.011 \\ \tiny (-0.072--0.037)} & \makecell{0.092 \\ \tiny (-0.060--0.258)} & \makecell{-0.008 \\ \tiny (-0.026--0.011)} \\
 & \makecell{4} & \makecell{0.673 \\ \tiny (0.642--0.700)} & \makecell{0.237 \\ \tiny (0.206--0.261)} & \makecell{0.336 \\ \tiny (0.272--0.398)} & \makecell{0.583 \\ \tiny (0.548--0.615)} & \makecell{0.475 \\ \tiny (0.304--0.616)} & \makecell{0.452 \\ \tiny (0.383--0.520)} & \makecell{0.254 \\ \tiny (0.216--0.295)} & \makecell{0.015 \\ \tiny (-0.075--0.130)} & \makecell{0.286 \\ \tiny (0.233--0.333)} & \makecell{\textbf{0.376} \\ \tiny (0.202--0.522)} & \makecell{0.128 \\ \tiny (0.115--0.142)} \\
TabPFN & \makecell{6} & \makecell{0.801 \\ \tiny (0.785--0.822)} & \makecell{0.652 \\ \tiny (0.638--0.669)} & \makecell{0.375 \\ \tiny (0.313--0.434)} & \makecell{0.504 \\ \tiny (0.468--0.537)} & \makecell{0.685 \\ \tiny (0.611--0.757)} & \makecell{0.478 \\ \tiny (0.393--0.549)} & \makecell{0.337 \\ \tiny (0.297--0.392)} & \makecell{0.286 \\ \tiny (0.203--0.345)} & \makecell{0.333 \\ \tiny (0.276--0.379)} & \makecell{0.316 \\ \tiny (0.154--0.460)} & \makecell{0.411 \\ \tiny (0.387--0.429)} \\
 & \makecell{8} & \makecell{\underline{0.926} \\ \tiny (0.917--0.935)} & \makecell{0.681 \\ \tiny (0.665--0.694)} & \makecell{0.435 \\ \tiny (0.375--0.497)} & \makecell{0.557 \\ \tiny (0.524--0.587)} & \makecell{\underline{0.691} \\ \tiny (0.636--0.771)} & \makecell{0.586 \\ \tiny (0.535--0.649)} & \makecell{0.310 \\ \tiny (0.267--0.350)} & \makecell{0.374 \\ \tiny (0.304--0.422)} & \makecell{0.411 \\ \tiny (0.357--0.457)} & \makecell{0.308 \\ \tiny (0.165--0.429)} & \makecell{\underline{0.619} \\ \tiny (0.603--0.636)} \\
 & \makecell{10} & \makecell{\textbf{0.960} \\ \tiny (0.954--0.965)} & \makecell{\textbf{0.716} \\ \tiny (0.701--0.730)} & \makecell{0.440 \\ \tiny (0.376--0.497)} & \makecell{\textbf{0.805} \\ \tiny (0.790--0.816)} & \makecell{0.682 \\ \tiny (0.587--0.747)} & \makecell{\textbf{0.672} \\ \tiny (0.609--0.725)} & \makecell{0.348 \\ \tiny (0.302--0.381)} & \makecell{\textbf{0.446} \\ \tiny (0.395--0.510)} & \makecell{0.456 \\ \tiny (0.416--0.494)} & \makecell{\underline{0.330} \\ \tiny (0.185--0.441)} & \makecell{\textbf{0.731} \\ \tiny (0.724--0.740)} \\
\bottomrule
\end{tabular}}
\end{table*}

\section{Dataset Characterization and Scalability Analyses}\label{sec:appendix_characterization_scalability}

This section provides the full dataset characterization and scalability analyses referenced in Section~\ref{sec:experiments}.

\subsection{Dataset Characterization and Performance Predictors}\label{sec:dataset_characterization}

\begin{figure}[t!]
    \centering
    \includegraphics[width=0.9\linewidth]{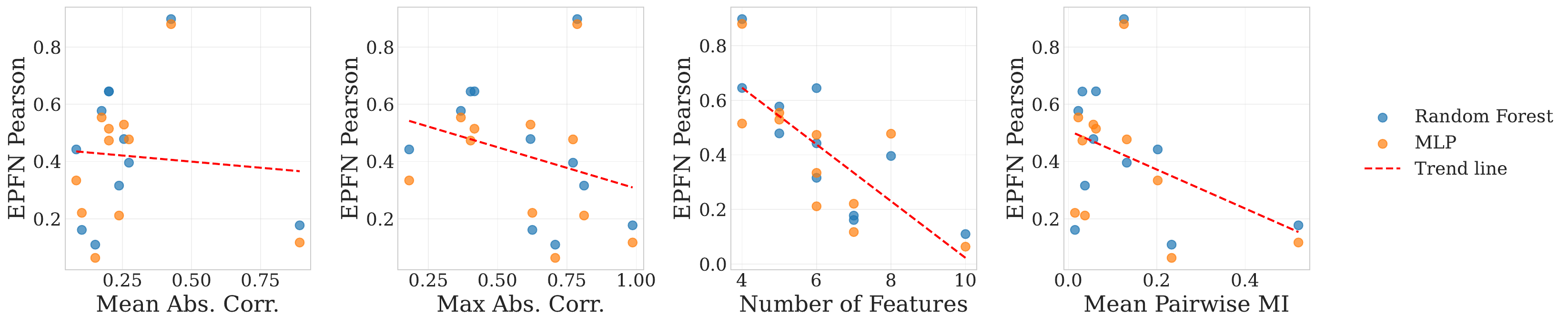}
    \caption{Dataset characterization analysis. Each point represents one evaluation dataset/base-model combination. Zero-shot ExplainerPFN performance (Pearson $r$ with SHAP) is plotted against (a) mean absolute pairwise correlation (Mean Abs.~Corr.), (b) maximum absolute pairwise correlation (Max Abs.~Corr.), (c) number of features, and (d) interaction depth (Mean Pairwise MI). A dashed red trend line is fitted to all points in each subplot. Dimensionality shows the strongest negative relationship with performance.}
    \label{fig:dataset_characterization}
\end{figure}

To understand when zero-shot attributions are most reliable, we characterize the 11 UCI evaluation datasets along structural and statistical dimensions and correlate each characteristic with ExplainerPFN's zero-shot Pearson correlation. We compute mean absolute pairwise correlation, maximum absolute pairwise correlation, average pairwise interaction depth via normalized mutual information (MI), and dimensionality $m$ for each dataset. Correlating these statistics with zero-shot performance reveals that \textbf{dimensionality is the dominant predictor of degradation}: the number of features exhibits a strong negative Spearman correlation of $-0.827$ with zero-shot Pearson performance (Figure~\ref{fig:dataset_characterization}), followed by interaction depth ($r_s = -0.329$) and maximum absolute pairwise correlation ($r_s = -0.360$). Table~\ref{tab:dataset_characterization} summarizes per-dataset values for all four characteristics alongside ExplainerPFN performance.

Table~\ref{tab:dataset_characterization} provides per-dataset values for all four characteristics alongside ExplainerPFN performance averaged across base predictors.

\begin{table}[h]
    \centering
    \caption{Per-dataset characterization statistics and ExplainerPFN zero-shot performance (Pearson and Spearman correlation with SHAP, averaged across RF and MLP base predictors).}
    \label{tab:dataset_characterization}
    \vspace{0.5em}
    \footnotesize
\begin{tabular}{lrrrrrrr}
\toprule
Dataset & $m$ & Mean Abs. Corr. & Max Abs. Corr. & Interaction Depth & Pearson & Spearman \\
\midrule
BA & 4 & 0.426 & 0.787 & 0.126 & 0.889 & 0.922 \\
AB & 7 & 0.891 & 0.987 & 0.521 & 0.147 & 0.282 \\
AN & 6 & 0.082 & 0.180 & 0.202 & 0.388 & 0.302 \\
CC & 5 & 0.255 & 0.618 & 0.056 & 0.504 & 0.519 \\
EC & 8 & 0.273 & 0.772 & 0.132 & 0.437 & 0.458 \\
FL & 10 & 0.151 & 0.708 & 0.233 & 0.087 & 0.242 \\
GC & 7 & 0.102 & 0.625 & 0.015 & 0.191 & 0.169 \\
HE & 6 & 0.201 & 0.402 & 0.031 & 0.559 & 0.492 \\
HD & 5 & 0.174 & 0.367 & 0.022 & 0.566 & 0.464 \\
HP & 4 & 0.200 & 0.416 & 0.062 & 0.580 & 0.139 \\
TH & 6 & 0.237 & 0.812 & 0.037 & 0.264 & 0.402 \\
\bottomrule
\end{tabular}

\end{table}

These findings yield practical guidance for practitioners: zero-shot attributions from ExplainerPFN are most trustworthy on low-dimensional datasets ($m \lesssim 8$) with well-conditioned feature correlations. On higher-dimensional or ill-conditioned datasets, practitioners should treat zero-shot estimates as rough directional indicators rather than precise quantitative attributions, and should prefer few-shot surrogates when even a small number of reference SHAP values are available.

\subsection{Scalability to Higher Dimensionality}\label{sec:scalability}

\begin{figure}[t!]
    \centering
    \includegraphics[width=0.6\linewidth]{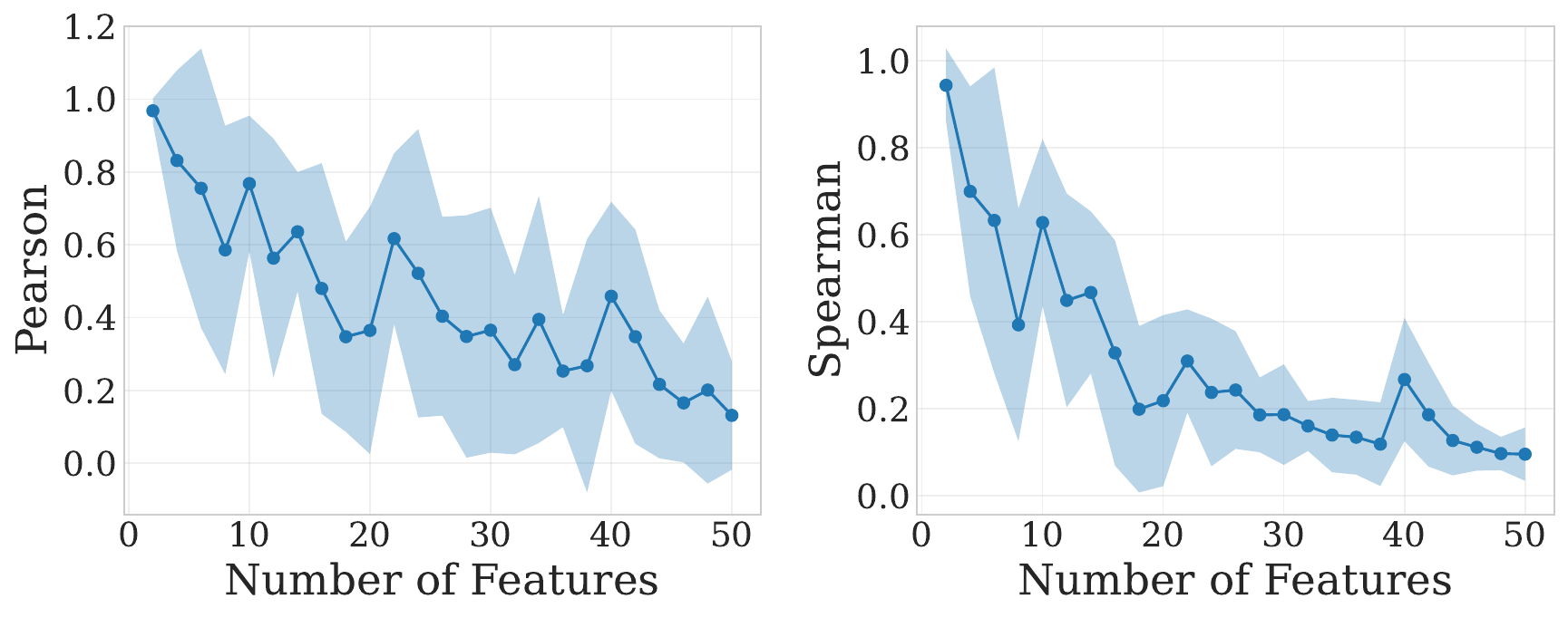}
    \caption{Scalability stress tests. ExplainerPFN zero-shot performance versus number of features on synthetic datasets ($n = 1{,}000$, 10 replicates per level). The left plot shows Pearson correlation; the right plot shows Spearman correlation. Solid lines show mean correlation; shaded regions show $\pm$1 standard deviation. Performance declines as dimensionality increases, with moderate variance across replicates. By $m = 50$, mean Pearson correlation remains small but positive ($\bar{r} = 0.13$).}
    \label{fig:scalability_dimensionality}
\end{figure}

We extend the dimensionality analysis beyond the $m \in [2, 20]$ range used in the ablation study (Figure~\ref{fig:ablation_combined}) by generating synthetic datasets with up to $m = 50$ features, holding interaction depth and sample size roughly constant at $n = 1{,}000$ observations. For each dimensionality level we run ten independent replicates to quantify variability. Figure~\ref{fig:scalability_dimensionality} shows the resulting Pearson and Spearman correlations between ExplainerPFN and exact Kernel SHAP.

Performance declines as dimensionality increases, though moderate variance across replicates (shaded regions in Figure~\ref{fig:scalability_dimensionality}) indicates that structural complexity drives the decline (and not merely dimensionality): some high-dimensional synthetic tasks retain moderately positive correlation while others collapse. By $m = 50$, mean Pearson correlation remains small but positive ($\bar{r} = 0.13$). Beyond this point, architectural modifications such as a larger context window, explicit positional embeddings for feature sets, or hierarchical attention over feature groups would likely be needed to maintain attribution fidelity. We leave these extensions to future work.

\section{Extended Case Study: ACS Public Coverage}\label{sec:appendix_acs_case_study}

This section extends the ACS Public Coverage case study in Section~\ref{sec:case_study} with per-instance visual comparisons. SHAP values were computed using the trained MLP (Section~\ref{sec:case_study}) with the empirical training distribution as background. We select one high-agreement and one low-agreement instance by per-instance Pearson correlation between SHAP and ExplainerPFN attributions.

\begin{figure}[h!]
    \centering
    \includegraphics[width=0.7\linewidth]{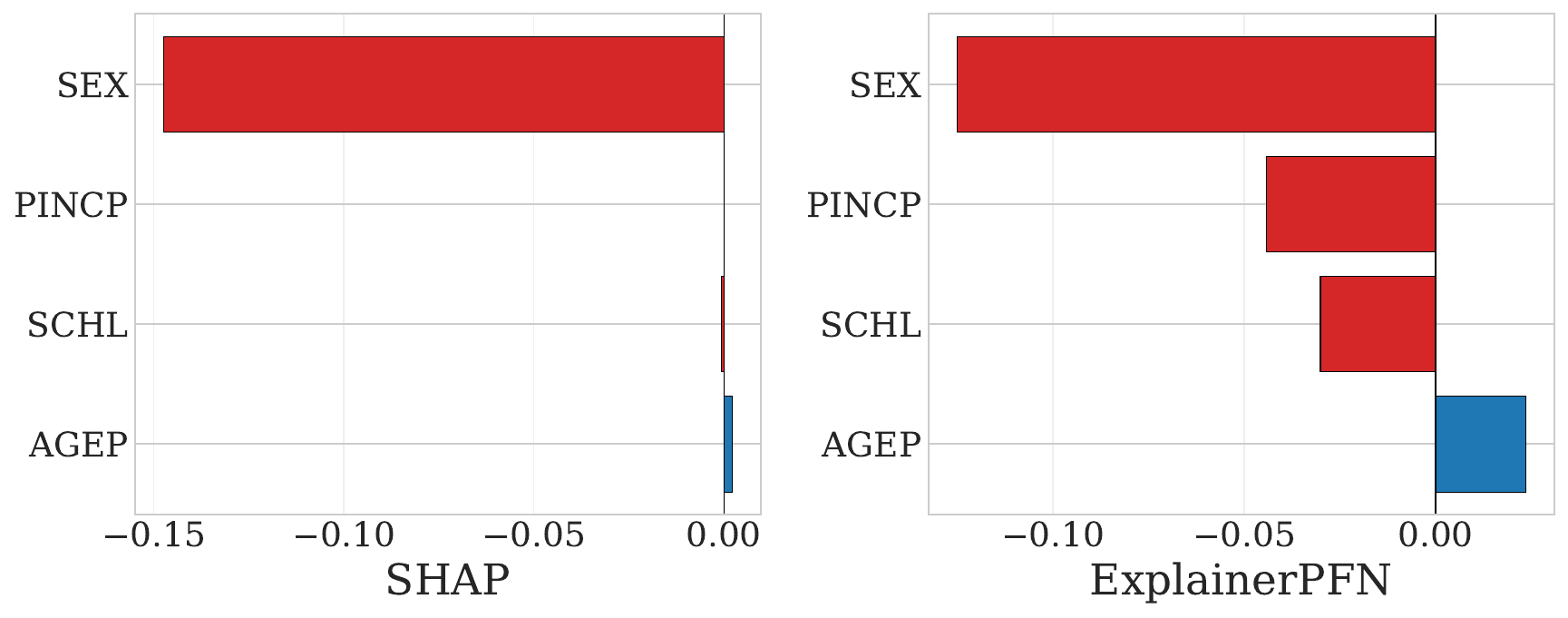}
    \caption{High-agreement instance between SHAP and ExplainerPFN on ACS Public Coverage. Feature values: AGEP=19, SCHL=16, PINCP=28000, SEX=2.}
    \label{fig:acs_waterfall_high}
\end{figure}

\begin{figure}[h!]
    \centering
    \includegraphics[width=0.7\linewidth]{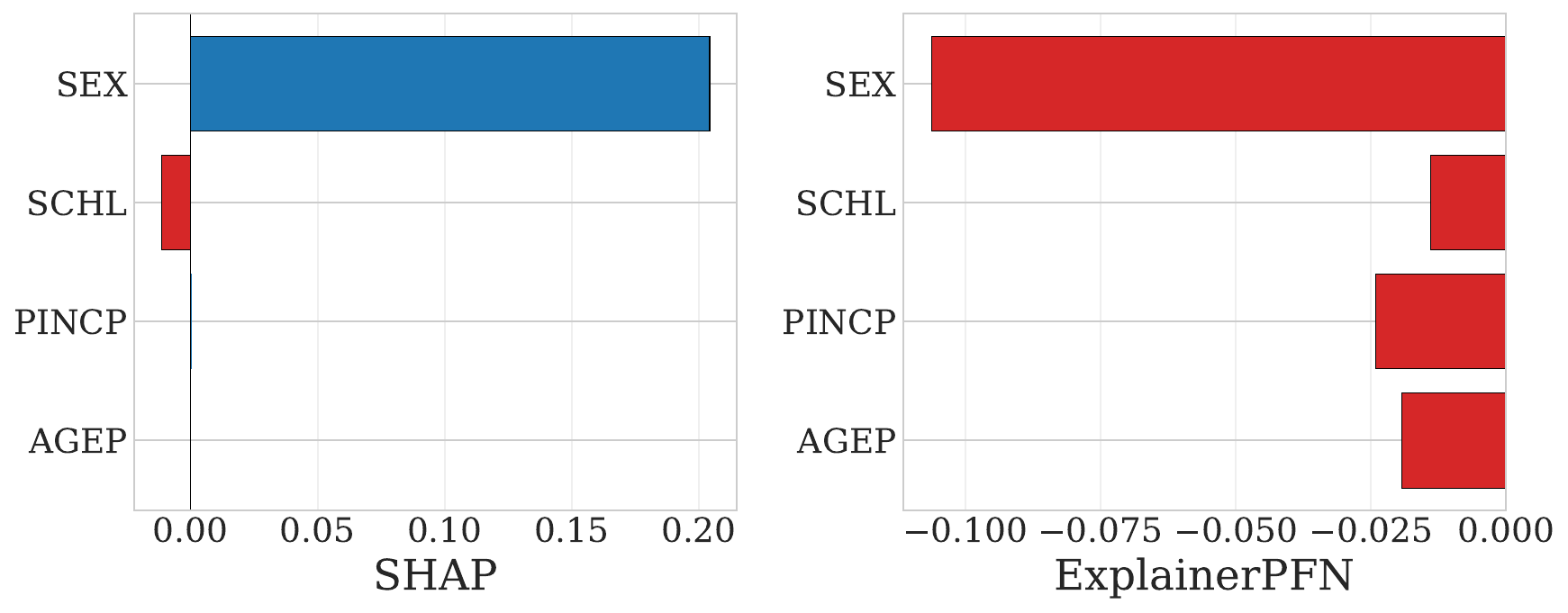}
    \caption{Low-agreement instance between SHAP and ExplainerPFN on ACS Public Coverage. Feature values: AGEP=35, SCHL=1, PINCP=30000, SEX=1.}
    \label{fig:acs_waterfall_low}
\end{figure}

Figure~\ref{fig:acs_waterfall_high} shows a high-agreement instance ($r = 0.887$): both methods agree on the dominant role of SEX (large negative importance), and three of the four features have the same directional sign. Figure~\ref{fig:acs_waterfall_low} shows a low-agreement instance ($r = -0.999$): SHAP and ExplainerPFN assign opposite signs for three of the four features, with the largest discrepancy on SEX, which SHAP identifies as strongly positive while ExplainerPFN strongly negative. This inversion illustrates how zero-shot attributions can diverge from model-specific SHAP values on individual instances even when aggregate trends align.

\newpage
\section*{NeurIPS Paper Checklist}

\begin{enumerate}

\item {\bf Claims}
    \item[] Question: Do the main claims made in the abstract and introduction accurately reflect the paper's contributions and scope?
    \item[] Answer: \answerYes{} 
    \item[] Justification: We provide the abstract and introduction reflecting and supporting the paper’s contribution and scope with detailed explanations.

\item {\bf Limitations}
    \item[] Question: Does the paper discuss the limitations of the work performed by the authors?
    \item[] Answer: \answerYes{} 
    \item[] Justification: See section~\ref{sec:conc} of the paper for a
      discussion of limitations and future work.

\item {\bf Theory assumptions and proofs}
    \item[] Question: For each theoretical result, does the paper provide the full set of assumptions and a complete (and correct) proof?
    \item[] Answer: \answerNA{} 
    \item[] Justification: Our paper does not claim theoretical results.

    \item {\bf Experimental result reproducibility}
    \item[] Question: Does the paper fully disclose all the information needed to reproduce the main experimental results of the paper to the extent that it affects the main claims and/or conclusions of the paper (regardless of whether the code and data are provided or not)?
    \item[] Answer: \answerYes{} 
    \item[] Justification: The paper provides detailed instructions for
      reproducing the main experimental results in
      Section~\ref{sec:experiments} and the supplemental material, including
      details about the datasets used, the model architecture, the training
      procedure, and the hyperparameters.

\item {\bf Open access to data and code}
    \item[] Question: Does the paper provide open access to the data and code, with sufficient instructions to faithfully reproduce the main experimental results, as described in supplemental material?
    \item[] Answer: \answerYes{} 
    \item[] Justification: All the code, data and model weights are available
      at a public URL provided in the first footnote of the paper.

\item {\bf Experimental setting/details}
    \item[] Question: Does the paper specify all the training and test details (e.g., data splits, hyperparameters, how they were chosen, type of optimizer) necessary to understand the results?
    \item[] Answer: \answerYes{} 
    \item[] Justification: The paper describes the experimental setting in
      detail in Section~\ref{sec:experiments} and the supplemental material.

\item {\bf Experiment statistical significance}
    \item[] Question: Does the paper report error bars suitably and correctly defined or other appropriate information about the statistical significance of the experiments?
    \item[] Answer: \answerYes{} 
    \item[] Justification: See Appendix~\ref{sec:appendix_full_tables}.

\item {\bf Experiments compute resources}
    \item[] Question: For each experiment, does the paper provide sufficient information on the computer resources (type of compute workers, memory, time of execution) needed to reproduce the experiments?
    \item[] Answer: \answerYes{} 
    \item[] Justification: The paper provides detailed information about the compute resources used for the experiments in Section~\ref{sec:experiments}.
    
\item {\bf Code of ethics}
    \item[] Question: Does the research conducted in the paper conform, in every respect, with the NeurIPS Code of Ethics \url{https://neurips.cc/public/EthicsGuidelines}?
    \item[] Answer: \answerYes{} 
    \item[] Justification: No human subjects were involved in the research, and the societal impact of this work is discussed in the impact statement.

\item {\bf Broader impacts}
    \item[] Question: Does the paper discuss both potential positive societal impacts and negative societal impacts of the work performed?
    \item[] Answer: \answerYes{} 
    \item[] Justification: The paper discusses potential positive and negative societal impacts in the impact statement.
    
\item {\bf Safeguards}
    \item[] Question: Does the paper describe safeguards that have been put in place for responsible release of data or models that have a high risk for misuse (e.g., pre-trained language models, image generators, or scraped datasets)?
    \item[] Answer: \answerNA{} 
    \item[] Justification: Not applicable.

\item {\bf Licenses for existing assets}
    \item[] Question: Are the creators or original owners of assets (e.g., code, data, models), used in the paper, properly credited and are the license and terms of use explicitly mentioned and properly respected?
    \item[] Answer: \answerYes{} 
    \item[] Justification: All the assets used to conduct the research are properly referenced and the source code and datasets are released under an MIT license.

\item {\bf New assets}
    \item[] Question: Are new assets introduced in the paper well documented and is the documentation provided alongside the assets?
    \item[] Answer: \answerYes{} 
    \item[] Justification: All the code base used to develop this paper is fully documented.

\item {\bf Crowdsourcing and research with human subjects}
    \item[] Question: For crowdsourcing experiments and research with human subjects, does the paper include the full text of instructions given to participants and screenshots, if applicable, as well as details about compensation (if any)? 
    \item[] Answer: \answerNA{} 
    \item[] Justification: Human subjects were not involved in the research.

\item {\bf Institutional review board (IRB) approvals or equivalent for research with human subjects}
    \item[] Question: Does the paper describe potential risks incurred by study participants, whether such risks were disclosed to the subjects, and whether Institutional Review Board (IRB) approvals (or an equivalent approval/review based on the requirements of your country or institution) were obtained?
    \item[] Answer: \answerNA{} 
    \item[] Justification: Human subjects were not involved in the research.

\item {\bf Declaration of LLM usage}
    \item[] Question: Does the paper describe the usage of LLMs if it is an important, original, or non-standard component of the core methods in this research? Note that if the LLM is used only for writing, editing, or formatting purposes and does \emph{not} impact the core methodology, scientific rigor, or originality of the research, declaration is not required.
    \item[] Answer: \answerNA{} 
    \item[] Justification: LLMs were used for editing and formatting purposes,
      as well as auditing the paper for clarity and consistency, as well as the
      code base for potential issues. However, LLMs were not used in the core
      method development of this research.

\end{enumerate}

\end{document}